%% file: DSN2019.tex
\documentclass[conference, compsocconf]{IEEEtran}
\input{packages.tex}

\newcommand*\circled[1]{\tikz[baseline=(char.base)]{
        \node[shape=circle,draw,inner sep=1.5pt, Maroon, fill=Maroon] (char)
               {\color{white}\scriptsize\textbf{#1}};}%
        }
        
\newcommand{\dstop}{d_{\text{stop}}}
\newcommand{\dsafe}{d_{\text{safe}}}
\DeclareMathOperator*{\Prob}{Pr}
\DeclareMathOperator*{\Prdo}{do}
\newcommand\given[1][]{\:#1\vert\:}
\DeclareMathOperator*{\argmax}{arg\,max}

\newcommand{\toolx}{DriveFI\xspace}
\newcommand{\apollo}{Apollo\xspace}
\newcommand{\dw}{DriveAV\xspace}

\begin{document}

\title{ML-based Fault Injection for Autonomous Vehicles:\\[-4pt] {\LARGE A Case for Bayesian Fault Injection}
}

\author{\IEEEauthorblockN{
    {Saurabh Jha}\IEEEauthorrefmark{1},
    {Subho S. Banerjee}\IEEEauthorrefmark{1},
    {Timothy Tsai}\IEEEauthorrefmark{2},
    {Siva K. S. Hari}\IEEEauthorrefmark{2},
    {Michael B. Sullivan}\IEEEauthorrefmark{2},\\
    {Zbigniew T. Kalbarczyk}\IEEEauthorrefmark{1},
    {Stephen W. Keckler}\IEEEauthorrefmark{2}
    and
    {Ravishankar K. Iyer}\IEEEauthorrefmark{1}
    }
    \IEEEauthorblockA{\IEEEauthorrefmark{1}{University of Illinois at Urbana-Champaign, Urbana-Champaign, IL 61801, USA.}}
    \IEEEauthorblockA{\IEEEauthorrefmark{2}{NVIDIA Corporation,  Santa Clara, CA 94086, USA.}}
}
\maketitle

\input{abstract}

\input{01000-introduction}
\input{02000-overview.tex}
\input{03000-method-pgm}
\input{04000-av.tex}
\input{05000-drivefi}

\input{06000-results}
\input{06010-GPU-FI.tex}

\input{06100-SLI.tex}

\input{06200-pgm.tex}

\input{08000-related}
\input{09000-conclusions}

\input{ack}

{
    \bibliographystyle{IEEEtran}
    \bibliography{references}
}

\end{document}

%% file: packages.tex
\usepackage[utf8]{inputenc}
\usepackage[english]{babel}
\PassOptionsToPackage{bookmarks=false}{hyperref}
\usepackage{multirow, multicol, booktabs, tabulary, tabu, longtable, array, varwidth}
\usepackage{placeins, lipsum}
\usepackage[flushleft]{threeparttable}

\usepackage[hyphens]{url}
\usepackage[labelfont=bf]{caption}
\usepackage{cite}
\usepackage{hyperref}
\usepackage[dvipsnames, table]{xcolor}
\hypersetup{
    linkcolor  = violet!85!black,
    citecolor  = magenta!85!black,
    urlcolor   = blue!85!black,
    colorlinks = true,
    breaklinks = true
}

\usepackage{enumerate}

\usepackage[inline]{enumitem}

\usepackage{amsmath, amsfonts, amssymb, amsthm, nicefrac}
\usepackage{siunitx}
\theoremstyle{definition}
\newtheorem{definition}{Definition}

\usepackage{listings}

\usepackage[scaled=0.82]{beramono}

\usepackage[]{cleveref} %
\crefname{appsec}{Appendix}{Appendices}
\crefformat{section}{\S#2#1#3}
\crefformat{subsection}{\S#2#1#3}
\crefformat{subsubsection}{\S#2#1#3}
\crefformat{equation}{(#2#1#3)}
\crefrangeformat{equation}{(#3#1#4--#5#2#6)}
\crefmultiformat{equation}{(#2#1#3)}{ and~(#2#1#3)}{, (#2#1#3)}{ and~(#2#1#3)}
\crefformat{figure}{Fig.~#2#1#3}
\crefrangeformat{figure}{Figs. #3#1#4--#5#2#6}
\crefmultiformat{figure}{Figs.~#2#1#3}{ and~#2#1#3}{, #2#1#3}{ and~#2#1#3}
\crefname{algocf}{Alg.}{Algs.}
\Crefname{algocf}{Algorithm}{Algorithms}
\crefformat{table}{Table~#2#1#3}
\crefrangeformat{table}{Tables~#3#1#4--#5#2#6}
\crefmultiformat{table}{Tables~#2#1#3}{ and~#2#1#3}{, #2#1#3}{ and~#2#1#3}

\usepackage{textcomp}
\usepackage[shortcuts,acronym]{glossaries}
\usepackage{soul, footnote, xargs}
\usepackage[colorinlistoftodos,prependcaption,textsize=tiny]{todonotes}
\usepackage{balance}

\usepackage{graphicx}
\usepackage{subcaption}
\usepackage{tikz, epstopdf, stfloats, bbding, capt-of}
\usepackage{algorithmic}
\usepackage[ruled, linesnumbered]{algorithm2e}

\SetCommentSty{mycommfont}
\usepackage{microtype}
\usepackage[T1]{fontenc}

\def\baselinestretch{0.92}

\makeatletter
\newcommand\subparagraph{%
    \@startsection{subparagraph}{5}
    {\parindent}
    {3.25ex \@plus 1ex \@minus .2ex}
    {-1em}
    {\normalfont\normalsize\bfseries}}
\makeatother
\usepackage[compact]{titlesec}
\let\subparagraph\relax %
\titlespacing\section{0pt}{6pt plus 4pt minus 2pt}{2pt plus 2pt minus 2pt}
\titlespacing{\subsection}{0pt}{4pt plus 2pt minus 1pt}{2pt plus 1pt minus 1pt}
\titlespacing{\subsubsection}{0pt}{4pt plus 2pt minus 1pt}{2pt plus 1pt minus 1pt}
\usepackage{etoolbox}
\makeatletter
\patchcmd{\ttlh@hang}{\parindent\z@}{\parindent\z@\leavevmode}{}{}
\patchcmd{\ttlh@hang}{\noindent}{}{}{}
\makeatother

\usepackage{soul}
\sethlcolor{black}
\makeatletter
\newif\if@blind
\@blindtrue %
\if@blind \sethlcolor{black}\else
   
\fi
\newcommand{\spacefigure}{\vspace{-2mm}}

%% file: abstract.tex
\begin{abstract}
	
The safety and resilience of fully autonomous vehicles (AVs) are of significant concern, as
exemplified by several headline-making accidents. While AV development today involves
verification, validation, and testing, end-to-end assessment of AV systems under accidental faults in
realistic driving scenarios has been largely unexplored. This paper presents DriveFI, a machine
learning-based fault injection engine, which can mine situations and faults that maximally impact AV
safety, as demonstrated on two industry-grade AV technology stacks (from NVIDIA and Baidu). For
example, DriveFI found 561 safety-critical faults in less than 4 hours. In comparison, random
injection experiments executed over several weeks could not find any safety-critical faults.

\end{abstract}

\begin{IEEEkeywords}
Autonomous Vehicles; Fault Injection
\end{IEEEkeywords}

%% file: 01000-introduction.tex
\section{Introduction}\label{s:intro}
Autonomous vehicles (AVs) are complex systems that use artificial intelligence (AI) and machine
learning (ML) to integrate mechanical, electronic, and computing technologies to make real-time
driving decisions. AI enables AVs to navigate through complex environments while maintaining a
\emph{safety envelope}~\cite{erlien2015shared,suh2016design} that is continuously measured and
quantified by onboard sensors (e.g., camera, LiDAR, RADAR)~\cite{driveav,apollo,openpilot}. Clearly,
the safety and resilience of AVs are of significant concern, as exemplified by several
headline-making AV crashes~\cite{teslaCrash,uberCrash}, as well as prior work characterizing
AV resilience during road tests~\cite{banerjee2018hands}. Hence there is a compelling need for a
comprehensive assessment of AV technology.

AV development today involves verification~\cite{Fan2017,DBLP:conf/tls/ClarkeG87,
DBLP:conf/popl/ClarkeES83,DBLP:conf/ftcs/BitnerJAAF94}, validation~\cite{shen1998native}, and
testing~\cite{DBLP:conf/iccad/RoyNPAS88,DBLP:conf/iccad/HamzaogluP00} as well as other forms of
assessment throughout the life cycle. However, assessment of these systems in realistic
execution environments, especially because of the occurrence of random faults, has been challenging. Fault
injection (FI) is a well-established method for testing the resilience and error-handling capabilities
of computing and cyber-physical systems~\cite{Hsueh1997} under faults. FI-based assessment of AVs
presents a unique challenge not only because of AV's complexity but also because of the centrality of
AI in a free-flowing operational environment~\cite{fraade2018measuring}.  Also, AVs represent a
complex integration of software~\cite{AVCode} and hardware technologies~\cite{AVHPC} that have been
shown to be vulnerable to hardware and software errors (e.g.,
SEUs~\cite{Esmaeilzadeh2011, karnik2004characterization},
\emph{Heisenbugs}~\cite{musuvathi2008finding}). Future trends of increasing code complexity and
shrinking feature sizes will only exacerbate the problem.

This paper presents \emph{DriveFI}, an intelligent FI framework for AVs that addresses the above challenge
by identifying hazardous situations that can lead to collisions and accidents. DriveFI includes
\begin{enumerate*}[label=(\alph*)]
    \item an FI engine that can modify the software and hardware states of an autonomous driving 
    system (ADS) to simulate the occurrence of faults, and
    \item an ML-based fault selection engine, which we call \emph{Bayesian fault injection}, that
	can find the situations and faults that are most likely to lead to violations of safety
	conditions.
\end{enumerate*} 
In contrast, traditional FI techniques~\cite{Hsueh1997} often do not focus on safety violations, and
in practice have low manifestation rates and require enormous amounts of time under
test~\cite{hari2017sassifi,Li2017}. Note that given a fault model, DriveFI can also perform
random FI to obtain a baseline.

\textbf{Contributions.}
DriveFI's Bayesian FI framework is able to find safety-critical situations and faults through causal and
counter-factual reasoning about the behavior of the ADS under a fault. It does so by
\begin{enumerate*}[label=(\alph*)]
    \item \emph{integrating domain knowledge} in the form of vehicle kinematics and AV architecture,
    \item \emph{modeling safety} based on lateral and longitudinal stopping distance, and
    \item using \emph{realistic fault models} to mimic soft errors and software errors.
\end{enumerate*}
Items (a), (b), and (c) are integrated into a \emph{Bayesian network} (BN). BNs provide a favorable
formalism in which to model the propagation of faults across AV system components with an
interpretable model. The model, together with fault injection results, can be used to design and
assess the safety of AVs. Further, BNs enable rapid probabilistic inference, which allows DriveFI to
quickly find safety-critical faults. The Bayesian FI framework can be extended to  other safety-critical systems (e.g., surgical robots). The framework requires specification of the safety constraints and the system software architecture to model causal relationship between the system sub-components.
We demonstrate the capabilities and generality of this
approach on two industry-grade, level-4 ADSs~\cite{safetyVision}: DriveAV~\cite{driveav} (a
proprietary ADS from NVIDIA) and Apollo 3.0~\cite{apollo} (an open-source ADS from Baidu). 

\textbf{Results.}
We use three fault models:
\begin{enumerate*}[label=(\alph*)]
    \item random and uniform faults in non-ECC-protected processor structures,
    \item random and uniform faults in ADS software module outputs (corrupted with min or max values), and
    \item faults in which ADS module outputs are corrupted with Bayesian FI.
\end{enumerate*}
The major results of our injection campaigns include: 
\begin{itemize}[noitemsep,nolistsep,leftmargin=*] 
    \item Using fault model (b) we compiled a list of 98,400 faults. An exhaustive
	evaluation of all 98,400 faults in our simulated driving scenarios would have taken 615 days. In
	comparison, our Bayesian FI was able to find 561 faults that maximally impact AV safety in less
	than 4 hours. Thus, Bayesian FI achieves $3690\times$ acceleration.
	Two cases found by Bayesian FI are described in \cref{sec:case_studies}; one, in particular,
	mimics the Tesla vehicle crash~\cite{teslaCrash}.

    \item Bayesian FI is able to find critical faults and scenes that led to safety hazards. 
    \begin{enumerate*}[label=(\alph*)]
        \item Out of the 561 identified faults, 460 manifested as safety hazards.
        \item These 460 faults were found to be associated with 68 safety-critical scenes\footnote{A scene is represented by one camera frame.} (out of 7200 scenes).

    \end{enumerate*}

	\item In comparison, several weeks of 5000 random FI experiments did not result in discovery of
	a single safety hazard. Only 1.93\% of the single-bit injections led to silent-data corruption (SDC) that caused
	actuation errors. The ADS recovered from all of these errors without any safety violations. In 7.35\%
	of the FIs, kernel panics and hangs occurred.  It is expected that recovery from such faults can be done with the backup/redundant systems that are present in AVs today.
\end{itemize}
We believe that the mining of critical situations by Bayesian FI will have wider applicability beyond
our fault injections here. Combining results from a range of fault injection experiments to create a
library of situations will help manufacturers to develop rules and conditions for AV testing and safe driving.

\textbf{Putting DriveFI in Perspective.}
Early work studied the safety of AVs using system-theoretic
approaches~\cite{abdulkhaleq2017systematic, leveson2004new}. More recent studies have focused on the
resilience of constituent modules of an ADS (described in~\cref{sec:av_arch}), e.g., \cite{Li2017,
Pei2017, salami2018resilience, reagen2018ares}. Another line of work~\cite{jha2018avfi,
rubaiyat2018experimental} has used FI to study sensor-related resilience in AVs. In contrast to
DriveFI, none of the prior approaches have considered the resilience of modern end-to-end 
AI-driven systems that use industry-grade ADSs to mine faults that lead to safety hazards.

%% file: 02000-overview.tex
\begin{figure}[!t]
    \centering
    \includegraphics[width=\columnwidth]{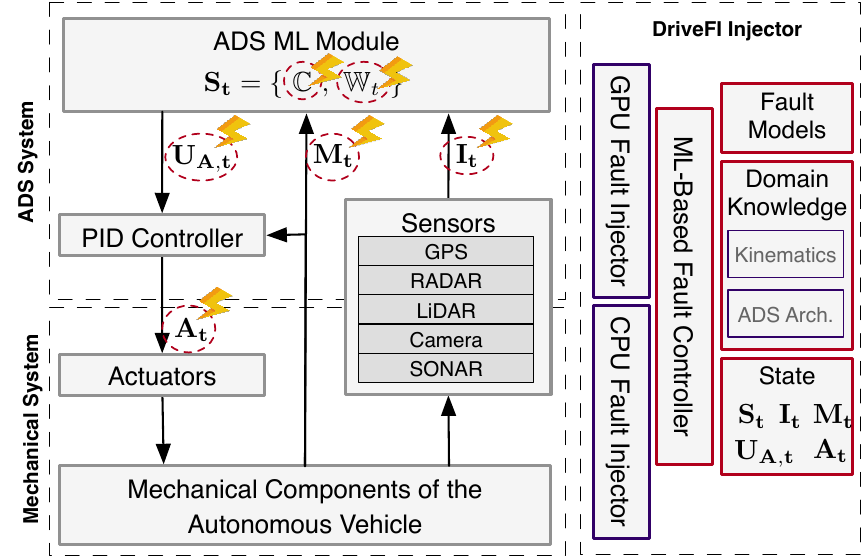}
    \caption{A high-level overview of the AV's autonomous and mechanical systems, and its interaction with DriveFI.}
    \label{fig:overview}
    \spacefigure 
    
\end{figure}

\section{Approach Overview} \label{sec:overview}

This section provides an overview of the AI-driven Bayesian FI approach advocated in this paper. We now
introduce the formalism that is used in the remainder of the paper.

\subsection{Autonomous Driving System}
\cref{fig:overview} illustrates the basic control architecture of an AV (henceforth also referred to
as \emph{Ego Vehicle}, EV). It consists of mechanical components and actuators that are
controlled by an \emph{ADS}, which represents the computational (hardware and software) component of
the AV. At every instant in time, $t$, the ADS system takes input from sensors $\mathbf{I_t}$ (e.g.,
cameras, LiDAR, GPS), takes inertial measurements $\mathbf{M_t}$ from the mechanical components
(e.g., velocity $v_t$, acceleration $a_t$), and infers actuation commands $\mathbf{A_t}$ (e.g.,
throttle $\zeta$, brake $b$, steering angle $\phi$). For clarity, we further subdivide the ADS into
two components:
\begin{enumerate*}[label=(\alph*)]
    \item an ML module (responsible for perception and planning) that takes as inputs 
    $\mathbf{I_t}$ and $\mathbf{M_t}$ and produces raw-actuation commands $\mathbf{U_{A, t}}$, and
    \item a PID controller~\cite{aastrom1995pid} that is responsible for smoothing the output 
    $\mathbf{U_{A, t}}$ to produce $\mathbf{A_t}$.
\end{enumerate*}
The PID controller ensures that the AV does not make any sudden changes in  $\mathbf{A_t}$. The ADS
ML module has an instantaneous state $\mathbf{S_t}$ that consists of configuration parameters $\mathbb{C}$
(e.g., neural network weights to perceive input camera data) and a \emph{world model} $\mathbb{W}_t$,
which maintains and tracks the trajectories of all static objects (e.g., lane markings) and dynamic objects (e.g., other vehicles) perceived by the ADS.

\begin{figure}[!t]
    \centering
    \includegraphics[width=0.8\columnwidth]{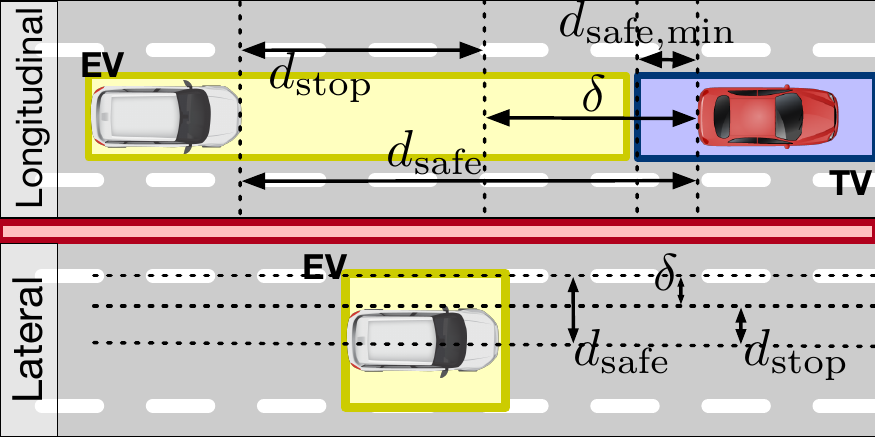}
    \caption{Definition of $\dstop$, $\dsafe$, and $\delta$ for lateral and longitudinal movement of the car. Non-AV vehicles are labeled as \emph{target vehicles} (TV).}
    \label{fig:safety}
    \spacefigure
\end{figure}

\subsection{Safety}
We define the instantaneous safety criteria of an AV in terms of the longitudinal (i.e., direction of
motion of the vehicle) and lateral (i.e., perpendicular to the direction of the vehicle motion)
Cartesian-distance travelled by the AV (see \cref{fig:safety}). Those criteria form a ``primal''
definition of safety based on collision avoidance, which can be extended with other notions of
safety, e.g., using traffic rules. The extended notions of safety are not considered in this paper,
as they can be nuanced based on the laws of the geographic regions in which they are applied.

\begin{definition}
    The \emph{stopping distance} $\dstop$ is defined as the maximum distance the vehicle will travel
	before coming to a complete stop while the maximum comfortable deceleration
	$a_{\text{max}}$ is being applied.
\end{definition}

\begin{definition}
    The \emph{safety envelope} $\dsafe$~\cite{erlien2015shared,suh2016design} of an AV is defined 
    as the maximum distance an AV can travel without colliding with any static or dynamic object.
\end{definition}

A safety envelope is used to ensure (through constraints on $\mathbf{U_{A, t}}$) that the vehicle
trajectory is collision-free. Production ADSs use techniques such as those in ~\cite{erlien2013safe,
anderson2012constraint} to estimate vehicle and object trajectories, thereby computing $\dsafe$
whenever an actuation command is sent to the mechanical components of the vehicle. These ADSs
generally set a minimum value of $\dsafe$ (i.e., $d_{\text{safe}, \text{min}}$) to ensure that a
human passenger is never uncomfortable about approaching obstacles.

\begin{definition}
	The \emph{safety potential} $\delta$ is defined as $\delta = \dsafe - \dstop$. An AV is defined
	to be in a \emph{safe state} when $\delta > 0$ in both lateral and longitudinal
	directions.\footnote{We use the shorthand $\delta > 0$ to mean both lateral and longitudinal
	$\delta$s.}
\end{definition}

\subsection{Fault Injection}
The goal of DriveFI is to test ADSs in the presence of faults to identify hazardous situations that
can lead to accidents (e.g., loss of property or life). To accomplish that goal, DriveFI includes
\begin{enumerate*}[label=(\alph*)]
    \item an FI engine that can modify the software and hardware states of the ADS to 
    simulate the occurrence of faults, and
    \item an ML-based fault selection engine that can find the faults and scenes that are most 
    likely to lead to violations of safety conditions and, hence, can be used to guide the fault 
    injection.
\end{enumerate*}
Taken together, these components of DriveFI can identify hazardous situations that lead to accidents
similar to the Tesla crash described later in this section.

\begin{figure*}[!t]
    \centering
    
    \begin{minipage}{.18\textwidth}
        \centering
        \includegraphics[width=\textwidth]{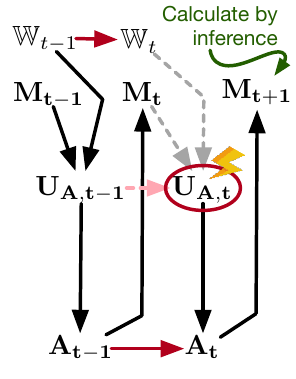}
        \caption{Bayesian FI.}
        \label{fig:bayesian_fault_injection}
    \end{minipage}%
    \hfill
    \begin{minipage}{.81\textwidth}
        \centering
        \includegraphics[width=\textwidth]{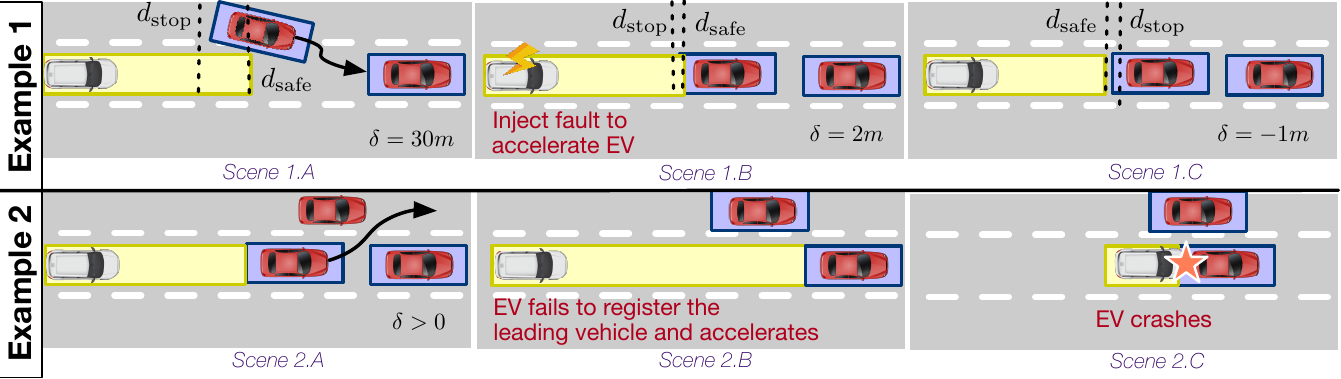}
        \caption{Example scenarios: (1) Targeted FI leads to hazardous conditions; (2) Real-world example with Tesla Autopilot that is similar to injected faults.}
        \label{fig:examples}
    \end{minipage}%
    \vspace{-1ex}
\end{figure*}

\textbf{Fault Model.}
We assume that faults injected in DriveFI can corrupt GPU architectural state. Memory and caches (of
both the CPUs and GPUs) are assumed to be protected with SECDED codes. Each injected fault is
characterized by its location (in this case, its dynamic instruction count) and the injected value.
The faults injected into the architectural states of these processors can manifest as \emph{errors}
in the inputs, outputs, and internal state of the ADS modules described above (i.e., $\mathbf{I_t}$,
$\mathbf{M_t}$, $\mathbf{S_t}$, $\mathbf{U_{A, t}}$ and $\mathbf{A_t}$). DriveFI can directly inject
errors into ADS outputs by corrupting the variables that store ADS outputs.
ADS software input/output variables are ultimately stored in different levels of storage hierarchies, e.g., registers or caches. Single- or multiple-bit faults cause corruption of variables when not masked in hardware~\cite{Avizienis2004}. Hence, faults are being injected into these memory units, but the variables are corrupted to emulate the faults. Therefore, our fault injectors target each element in the internal ADS software state ($S\_t$), sensor inputs ($I_t$), vehicle inertial measurements ($M_t$), and actuation commands ($U_t$, $A_t$), as shown in Fig.~\ref{fig:drivefi}. We define any error that causes
safety issues for the AV as \emph{hazardous}.
For simplicity and clarity, in the
remainder of the paper, we refer to both injected faults and errors as \emph{faults}.

To build a baseline for the ML-based targeted injections, we used DriveFI to perform random
injections into the GPU architectural state and ADS module outputs for two production ADS systems from
NVIDIA and Baidu. In contrast to prior work~\cite{Li2017, reagen2018ares}, which has reported
significant SDC rates (as high as 20\%) for the constituent deep-learning models (ConvNets that deal
with perception: object recognition and tracking) of the ADS system, we observed that random
injections rarely cause hazardous errors. These faults are masked because of the natural resilience
of the ADS stack, i.e.,
\begin{enumerate*}[label=(\alph*)]
    \item for production ADS systems that make real-time inferences at 60--100 Hz, transient faults have little chance to propagate to actuators before a new system state is recalculated;
    \item the ADS system architecture is inherently resilient, as it uses algorithms like extended 
    Kalman filtering~\cite{julier1997new} (for sensor fusion) and PID control (for output 
    smoothing); and
    \item not all driving scenes/frames are hazardous even under faults. Environmental conditions, 
    such as the presence of other objects on the streets, are fundamental in defining the safety 
    envelope.
\end{enumerate*}

\textbf{Bayesian Fault Injection.}
Consider a fault $f$ that changes the value of one of the aforementioned variables. The goal of the
ML-based fault injector is to find a \emph{critical situation} that is inherently safe (i.e.,
$\delta > 0$) and becomes unsafe after injection of fault $f$ (i.e., $\delta_{\Prdo(f)} \leq 0$). The
set of all faults $\mathbf{F_{\text{crit}}}$ in which that condition holds is defined as
\begin{equation}
    \mathbf{F_{\text{crit}}} = \left\{ f: \delta > 0 \land \hat\delta_{\Prdo(f)} \leq 0 \right\}.
    \label{eqn:fcrit}
\end{equation}

The solution to that problem requires causal and counter-factual reasoning about the behavior of the
ADS under a fault. DriveFI performs that reasoning by modeling the ADS system using a \emph{Bayesian network}
(BN; shown in \cref{fig:bayesian_fault_injection}), which can capture causal
relationships~\cite{Pearl2018}. The BN describes statistical relationships shown by black arrows
between the variables $\mathbb{W}_t$, $\mathbf{M_t}$, $\mathbf{U_{A, t}}$, and $\mathbf{A_t}$ at a
time $t$, as well as relationships shown by red arrows between the variables over time. The topology
of the BN is derived from the architecture of the ADS system. For example,
\cref{fig:bayesian_fault_injection} has the same graphical structure as \cref{fig:overview}. DriveFI
uses the BN to calculate the maximum likelihood estimate (MLE)\footnote{The estimated value of $x$
is denoted by $\hat x$.} of the value $\mathbf{\hat M_{t+1}}$ and then uses the MLE value to
calculate $\hat\delta_{\Prdo(f)}$ based on the kinematic model of the AV described later in
\cref{sec:bayesian_fault_injection}. We use probabilistic inference over the posterior distribution
of the BN to calculate
\begin{equation}
    \mathbf{\hat M_{t+1}} = \argmax_\mathbf{m} \Prob\left[\mathbf{M_{t+1}} = \mathbf{m} \given \Prdo(f)\right].
    \label{eqn:high_level_mle}
\end{equation}
The $\Prdo(\cdot)$ notation is based on the \emph{do-calculus} defined in~\cite{Pearl2018}. It marks
an FI action as an intervention in the BN model. It replaces certain probabilities with
constants and removes statistical conditional dependencies that are a target of the intervention
(i.e., dashed lines in \cref{fig:bayesian_fault_injection}), but preserves all other statistical
dependencies. We call this notion of counterfactual reasoning about the importance of a fault in
performing targeted injections \emph{Bayesian Fault Injection}.

\subsection{Case Studies}~\label{sec:case_studies}
To explain the need for a high-efficiency FI mechanism (such as our ML-based fault
injector), we discuss two examples of car accidents due to faults.

\textbf{Example 1: Hazardous Error.}
\cref{fig:examples} shows an example driving scenario in which a fault was injected into an
ADS through corruption of the throttle command (which was changed from 0.2 to 0.6). The injected
error led to an accident. We assume that (a) the ADS is running perception, planning, and control
inference at 30 Hz, and (b) all vehicles are running on a highway with a velocity of 33.5 m/s, which
is roughly the speed limit on U.S. freeways. In Scene 1A, the Ego vehicle (EV) was accelerating;
however, target vehicle TV\#1, operated by a human, initiated a lane change procedure, which
decreased the safety potential $delta$ from 20 m to 2 m as shown in ``Scene 1B.'' At that point, the
Bayesian fault injector injects a fault into the throttle command, causing the vehicle to
accelerate. The increase in acceleration caused the EV to become unsafe ($\delta < 0$), as shown in
``Scene 1C.'' The EV velocity is high enough that braking, even with $a_{\text{max}}$, is not able to prevent  an accident. This example shows that one needs a smart FI mechanism (such as our
Bayesian-based injector) that is able to inject a fault at a precise time instant based on a
run-time measurement of the safety potential to maximize the stress on the ADS and cause the EV to
crash. As we argue in later sections, it is impractical (or highly difficult) to achieve the
same objective using random FI.

\textbf{Example 2: Real-World Crash.}
Fig.~\ref{fig:examples} shows a real-world example of a fatal accident that was shown to have been
caused by a problem in Tesla Autopilot~\cite{teslaCrash}. In Scene 2A, the EV followed the lead
vehicle (TV\#1). A few seconds later, TV\#1 changed lanes (shown as Scene 2B); at that point,
Autopilot decided to accelerate in order to match the allowed highway speed. However, TV\#1 was
behind another vehicle (TV\#2), and the EV had no knowledge of TV\#2; it was too late for the EV to
recognize TV\#2 and slow down in time to avoid an accident. While this crash was attributed to a
design problem (i.e., delayed recognition) in the perception subsystem of the ADS, one can imagine
that a runtime fault (that delays perception of an object) could lead to the same fatal outcome. As
we show later, our Bayesian-based fault injector is able to recreate such scenarios.

%% file: 03000-method-pgm.tex
\begin{figure}[!t]
    \centering
    \includegraphics[]{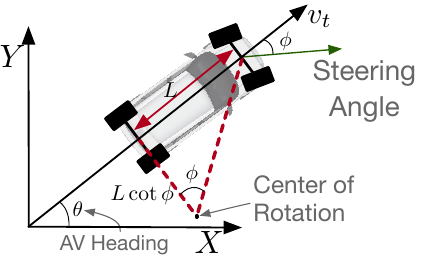}
    \caption{Orientation of the EV when in motion.}
    \label{fig:car_angle}
    \spacefigure
\end{figure}

\section{Bayesian Fault Injection}\label{sec:bayesian_fault_injection}
In this section, we describe in detail the formulation of the \emph{Bayesian Fault Injection}
approach.

\subsection{Kinematics-Based Model of Safety}\label{sec:kinematics}
Consider an EV moving in two-dimensional space as shown in \cref{fig:car_angle}. The vehicle at time
$t$ has an instantaneous position $(x_t, y_t)$, speed $v_t$, heading $\theta_t$, and steering angle
$\phi_t$. The equations of motion for the vehicle are
\begin{equation}
    \nicefrac{dx_t}{dt} = v_t \cos \theta_t; \nicefrac{dy_t}{dt} = v_t \sin \theta_t; \nicefrac{d\theta_t}{dt} = \nicefrac{(v_t \tan \phi_t)}{L},
    \label{eqn:EV_diff_eqns}
\end{equation}
where $L$ is the distance between the wheels of the EV~\cite{Lavalle2006}. Here $v_t$ and $\phi_t$
are determined by the control model for the EV. In our case, $v_t$ is defined based on the output of
the ADS $\mathbf{A_t}$, i.e., $v_t = f(\zeta_t, b_t, \phi_t)$.

Note that a more complete model of the EV motion might include other dynamics, e.g., \emph{sliding}
and \emph{skidding} of the EV's wheels. We do not add these complications to our model, as that
would require us to make additional assumptions that are beyond the scope of this paper, e.g., about the EV's
tires, road conditions, road banking, and weather. Similarly, we do not consider the 3-D motion of
the EV, as doing so would require further assumptions about the topology of the maps (e.g.,
elevation) in the FI campaign. Our approach can be extended to consider those additional
factors.

We can compute the maximum stopping distance $\dstop$ from \cref{eqn:EV_diff_eqns} by first 
computing the time $t_{\text{stop}}$ taken to bring the vehicle to a complete halt, i.e.,
\begin{equation}
    \frac{dx_t}{dt}\Bigr|_{\substack{t = t_{\text{stop}}}} = 0 \text{ and }
    \frac{dy_t}{dt}\Bigr|_{\substack{t = t_{\text{stop}}}} = 0.
    \label{eqn:tstop}
\end{equation}
$\dstop$ is then calculated as $[x_{t_{\text{stop}}}-x_0, y_{t_{\text{stop}}}-y_0]^T$, where $(x_0,
y_0)$ is the position of the EV at the beginning of the maneuver. Closed-form solutions to the
system of differential equations \cref{eqn:EV_diff_eqns,eqn:tstop} are intractable for arbitrary
control procedures (i.e., $v_t$ and $\phi_t$) and have to be solved by iterative numerical solution
methods like the \emph{Runge-Kutta methods}~\cite{devries1995first}.

\textbf{The Emergency Stop Maneuver.}
To simplify our analysis, we assume that the EV executes a special maneuver we call an \emph{emergency
stop} to bring the vehicle to a halt. This procedure is characterized by
\begin{equation}
    \frac{dv_t}{dt} = -a_{\text{max}} \text{ and } \frac{d\phi_t}{dt} = 0.
    \label{eqn:emergency_stop}
\end{equation}
That corresponds to the deceleration of the EV with the maximum deceleration to come to a halt. 
\cref{eqn:emergency_stop} reduces \cref{eqn:EV_diff_eqns} to
\begin{subequations}
    \begin{align}
        \nicefrac{d^2x_t}{dt^2} &= -a_{\text{max}} \sin \theta_t (\nicefrac{d\theta_t}{dt}) \\
        \nicefrac{d^2y_t}{dt^2} &= -a_{\text{max}} \cos \theta_t (\nicefrac{d\theta_t}{dt}) \\
        \frac{d\theta_t}{dt} &= \frac{(\sqrt{(\nicefrac{dx_t}{dt})^2 + (\nicefrac{dy_t}{dt})^2})}{L} \tan\phi_0,
    \end{align}
    \label{eqn:EV_diff_eqns_emergencystop}
\end{subequations}
where $\phi_0$ is the steering angle of the car at the beginning of the maneuver. DriveFI uses the
system of equations defined in  \cref{eqn:EV_diff_eqns_emergencystop,eqn:tstop} to find $\dstop$. We
use the shorthand $\mathcal{P}$ to denote the procedure (iterative numerical integration) used to
compute
\begin{equation}
    \dstop = \mathcal{P}(a_{\text{max}}, v_0, \theta_0, \phi_0, x_0, y_0)
    \label{eqn:dstop_diffeqn_estimator}
\end{equation}
from the above equations and the initial kinematic state of the EV (i.e., $v_0, \theta_0, \phi_0$,
$x_0$, $y_0$) at the start of the maneuver.

Recall from \cref{sec:overview} that $\delta = \dsafe - \dstop$ and that $\delta > 0$ defines the
safety of the EV. The $\dsafe$ value is assumed to be computed directly from the sensors of the EV.
It is the distance to the closest object (static or dynamic) in the longitudinal or lateral path of
the EV. As a result, $\dsafe$ changes with time, and it is updated at the sensor's (e.g., LiDAR's or
camera's) refresh rate. We include the boundaries of the lane in which the EV is travelling
(henceforth referred to as the \emph{Ego lane}) as a static object to be used in $\dsafe$ computations to
ensure that we capture lane violations as a safety hazard.

\textbf{Discretization.}
We convert the problem of solving \cref{eqn:dstop_diffeqn_estimator} from one that uses continuous
time to one that uses a discrete notion of time. Discrete time is a natural fit for the ADS, as the
control decisions are made at discrete steps that correspond to the sensors' sampling frequencies.
Hence we convert time $t$ to a discrete number $k \in \mathbb{N}$ such that $t = k\Delta t$, where
$\Delta t$ is the period of the sensor with the smallest sampling frequency. In the case of the
DriveFI injector in DriveWorks and Apollo, that is 7.5 Hz. However, our methodology is frequency-agnostic.

\subsection{ML Model}\label{sec:ml}
The goal of a targeted fault injector is to find situations in which $\delta > 0$, but under the
injection of a fault $f$ (which manifests as changes in the kinematic state of the EV) into the ADS
stack, $\delta_{\Prdo(f)} \leq 0$. A solution to that problem involves speculating forward in time
to after the fault has been injected, recomputing $\dstop$ under the fault, and then reevaluating the
safety criteria for the EV. We apply an ML algorithm, which has been trained as a predictor of the EV's kinematic
state, as the mechanism for speculation. We now describe the design of the model and its training
and inference.

\textbf{The Model.}
Consider a situation in which a fault is injected into the EV's ADS at time point $k$. We want to
estimate the value of $\dstop$ at time $k+1$ when the (corrupted) actuation commands of the previous
time step have been acted upon. As we showed in the previous section, we can do so using
\cref{eqn:dstop_diffeqn_estimator}. However, that would require knowing the values $x_{k+1}$,
$y_{k+1}$, $v_{k+1}$, $\theta_{k+1}$, and $\phi_{k+1}$ as the initial conditions to start the
emergency stop maneuver. DriveFI estimates those values based on a maximum likelihood estimation
over the posterior distribution of a probabilistic model that captures the components of the ADS.

\begin{figure}[!t]
    \centering
    \includegraphics[width=\columnwidth]{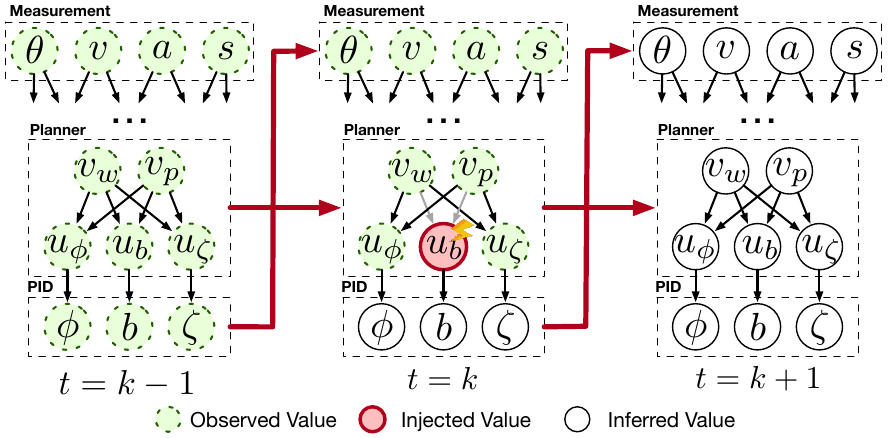}
    \caption{3-Temporal Bayesian Network modeling the ADS.}
    \label{fig:pgm-approach}
    \spacefigure
\end{figure}

DriveFI uses a Dynamic Bayesian Network (DBN)~\cite{koller1994towards}, specifically a  3-Temporal
Bayesian Network (TBN), i.e., a DBN unfolded thrice, to model $x_{k+1}$, $y_{k+1}$, $v_{k+1}$,
$\theta_{k+1}$, and $\phi_{k+1}$. This model is illustrated in \cref{fig:pgm-approach}. The core
idea of DBNs is to model each point in time with a static BN and to add temporal links from one
time-slice to the next (as shown by red arrows in \cref{fig:pgm-approach}). Usually all the
time-points have identical BN topologies and hyper-parameter settings. BNs are directed acyclic
graphs in which nodes represent random variables and arcs represent the causal connections among the
variables~\cite{pearl2014probabilistic}. Henceforthm we will refer to each random variable in the BN is henceforth referred to as
a \emph{node} to avoid confusion with the ADS variables. Each node $x$ is associated with a probability
table that provides conditional probability distributions (CPDs; $\Prob(x \given \pi(x)$) of a
node's possible value given the value of its parent nodes $\pi(x)$.

The 3-TBN model (see \cref{fig:pgm-approach}) is constructed based on the topological structure
shown in \cref{fig:overview}. A detailed version of this figure for the Apollo and DriveFI ADSs is
described in \cref{sec:av_arch} and shown in \cref{fig:appollo-arch}. The variables in each of the
ADS modules are connected in a parent-child fashion that reflects the data-flow in
\cref{fig:appollo-arch}. For example, the edges between $u_\zeta$ and $\zeta$ (in
\cref{fig:pgm-approach}) represent the CPD $\Prob(\zeta \given u_\zeta)$. This is an approximation
of the PID control for $\zeta$. Similarly, other components of the ADS are modeled based on their
input and output variables. We assume that the nodes in the 3-TBN are described by a CPD that has
the functional form
\[
    \Prob(x \given \pi(x)) = \mathcal{N}(\mathbf{\mu_x^T}\pi(x), \sigma_x)
\]
where $\mathcal{N}$ is the normal distribution with parameters $\mathbf{\mu_x}$ and $\sigma_x$ (for
each node $x$ in the network). That particular form of $\Prob(x \given \pi(x))$ is chosen because
\begin{enumerate*}[label=(\alph*)]
    \item it has numerical stability at small probability values, which are common when dealing with
	rare events like faults, and
    \item it simplifies the algorithm required to train the 3-TBN.
\end{enumerate*}

The use of the 3-TBN-based-modeling formalism is based on the implicit assumptions that
\begin{enumerate*}[label=(\alph*)]
    \item the EV state can be completely determined by its previous state and the observed software
	variables, and
    \item the transition parameters from one time step to another do not change with time, i.e.,
	the Markovian dynamic system is assumed to be homogeneous.
\end{enumerate*}

\textbf{Probabilistic Inference.}
The maximum likelihood estimate value $\hat v_{k+1}$ under a manifested fault $f$ (which corresponds
to setting the value of a variable in the model) is
\begin{equation}
    \hat v_{k+1} = \argmax_v \Prob\left( v_{k+1} = v \given \Prdo(f), \mathbf{O_k^{(f)}} \right).
    \label{eqn:mle_inf_with_obs}
\end{equation}
Given that we can execute a simulation of the EV under non-fault conditions, all variables that are
not children of the injected variable can be observed to have values from the correct run. These
``golden'' observations are labeled $\mathbf{O_k^{(f)}}$. \cref{eqn:mle_inf_with_obs} is solved
by first estimating the posterior distribution of $v_{k+1}$ by using \emph{Markov Chain Monte Carlo}
methods~\cite{koller1994towards} and then estimating the most likely value of $v_{k+1}$. A similar
procedure can be used to compute $\hat\theta_{k+1}$ and $\hat\phi_{k+1}$. The values of $\hat
x_{k+1}$ and $\hat y_{k+1}$ can then be computed using time-discretized versions of
\cref{eqn:EV_diff_eqns}. Finally, from \cref{eqn:dstop_diffeqn_estimator}, we get
\begin{equation}
    \hat \dstop = \mathcal{P}\left( a_\text{max}, \hat x_{k+1}, \hat y_{k+1}, \hat v_{k+1}, \hat \theta_{k+1}, \hat \phi_{k+1} \right).
    \label{eqn:dstop_estimator_final}
\end{equation}

\textbf{Training.}
The 3-TBN described above defines a probability distribution $\Prob(\mathbb{X}_{k-1}, \mathbb{X}_k,
\mathbb{X}_{k+1})$, where $\mathbb{X}_k = \mathbf{M_k} \cup \mathbf{S_k} \cup \mathbf{U_{A, k}} \cup
\mathbf{A_k}$. Via the BN formalism, $P(\mathbb{X}_{k-1}, \mathbb{X}_k, \mathbb{X}_{k+1})$ is
defined as
\[
    \Prob(\mathbb{X}_{k-1}, \mathbb{X}_k, \mathbb{X}_{k+1}) = \frac{1}{Z} \prod\displaylimits_{x \in \mathbb{X}_{k-1} \cup \mathbb{X}_k \cup \mathbb{X}_{k+1}} \Prob(x | \pi(x))
\]
where $Z$ is the partition function that normalizes $P$ to be a probability distribution. We use the
\emph{Expectation-Maximization algorithm}~\cite{dempster1977maximum} to compute
\begin{equation}
    \mathbf{\hat \mu}, \mathbf{\hat \sigma} = 
    \argmax_{\mathbf{\mu}, \mathbf{\sigma}} 
    E_{\mathbb{X} \given \mathcal{D}, \mathbf{\mu}, \mathbf{\sigma}}
    \left[ \log P(\mathbb{X} \given \mathbf{\mu}, \mathbf{\sigma}) \right]
    \label{eqn:em_training}
\end{equation}
where $\mathcal{D}$ refers to a training dataset that contains values of $\mathbb{X}_{k-1}$,
$\mathbb{X}_k$, and $\mathbb{X}_{k+1}$ under normal operation as well as during FIs. Here,
computation of $Z$ is intractable because of the combinatorially large size of $\mathbb{X}_{k-1}
\times \mathbb{X}_k \times \mathbb{X}_{k+1}$. However, \cref{eqn:mle_inf_with_obs} does not require the
computation of $Z$, as it is a common multiplicand to all values of the objective function.

\textbf{Training Data.}
The variables in $\mathbb{X}_k$ are measured by executing the ADS in several driving scenarios in a
simulator. We describe the setup of this simulator in \cref{sec:av_arch}. Simply capturing the data
under normal operation is not sufficient to capture abnormalities created in the ADS state because
of faults. Therefore, in addition to running driving scenarios without faults, we run the driving
scenarios while injecting random faults (i.e., the baseline described in \cref{sec:overview}) one at
a time. The FI campaign that corresponds to the training data is described in \cref{sec:drivefi}. We
recreate the process of injecting a fault into a uniformly randomly selected scene 20 to 50 times
for each fault. The reason for varying the number of faults is that some variables (such as
$\zeta$, $b$, and $\phi$) exhibit all possible values during simulated runs with no injections, while
others, such as stateful variables, simply do not vary naturally.

\begin{figure}[!t]
    \centering
    \includegraphics[width=\columnwidth]{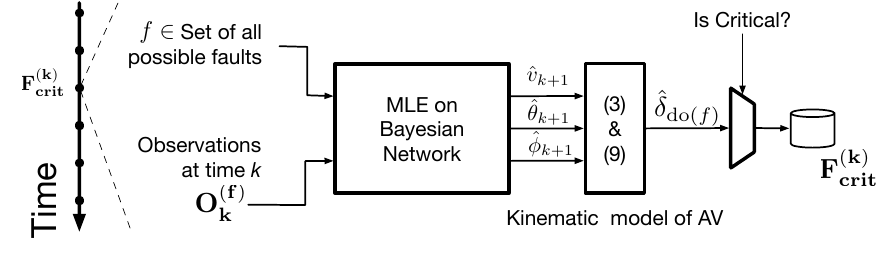}
    \caption{BN MLE inference is executed offline for every simulated time point to find the set of critical faults.}
    \label{fig:bayesian_fi_proc}
    \spacefigure
\end{figure}

\textbf{Fault Injection.}
The computation of $\mathbf{F_{\text{crit}}}$ (from \cref{eqn:fcrit}) is done offline for every
frame in every driving scene. The FI procedure executes as follows (see
\cref{fig:bayesian_fi_proc}):
\begin{itemize}[noitemsep,nolistsep,leftmargin=*]
    \item For each driving scenario, a non-fault-injected ``golden'' execution of the simulation is 
    performed. At each instant $k$, the variables in $\mathbb{X}_k$ are measured and stored.
    \item These ``golden'' values of $\mathbb{X}_k$ are stepped through with
	\cref{eqn:dstop_estimator_final} to build $\mathbf{F_{\text{crit}}^{(k)}}$ for every scene/frame,
	based on \cref{eqn:fcrit}.
    \item An FI campaign is carried out on the simulated EV to execute faults in 
    $\bigcup_k\mathbf{F_{\text{crit}}^{(k)}}$ one frame and one fault at a time.
\end{itemize}

%% file: 04000-av.tex
\section{The ADS Architecture \& Simulation}\label{sec:av_arch}

\subsection {AI Platform}
\begin{figure}[!t]
    \centering
    \includegraphics[width=\columnwidth]{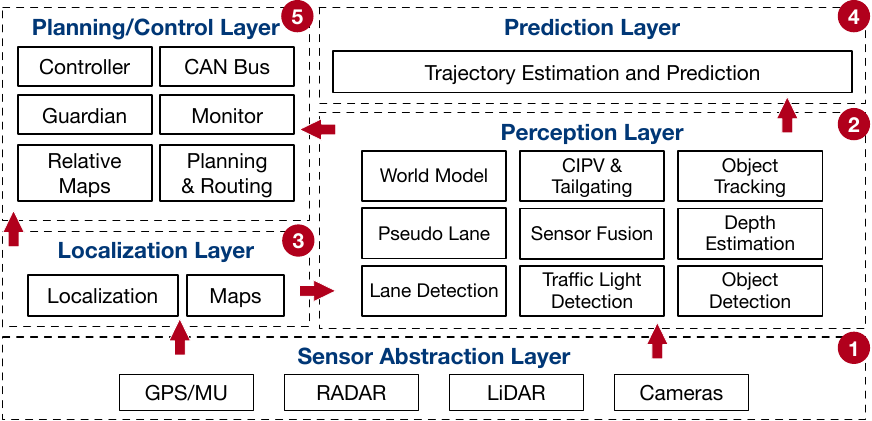}
    \caption{ADS architecture.}
    \label{fig:appollo-arch}
    \spacefigure
\end{figure}

An AV uses ADS technology to support and replace a human driver for
the tasks of controlling the vehicle's steering, acceleration, and monitoring of the
surrounding environment (e.g., other vehicles/pedestrians)~\cite{SAE_J3016_201609}.
The ADS architecture consists of five basic layers~\cite{apollo}, discussed
below:

\textbf{Sensor Abstraction Layer} (\circled{1} in \cref{fig:appollo-arch}): 
The sensor abstraction layer is responsible for preprocessing of input data, noise filtering, gains
control~\cite{watson1997model}, tone-mapping~\cite{debevec2002tone},
demosaicking~\cite{menon2011color}, and extraction of regions of interest, depending on the sensor type.
An ADS supports a wide range of sensors, such as Global Positioning System (GPS), Inertial
measurement unit (IMU), sonar, RADAR, LiDAR, and camera sensors. Our experiments only use two cameras
(fitted at the top and front of the vehicle) and one LiDAR.

\textbf{Perception Layer} (\circled{2} in \cref{fig:appollo-arch}):
The sensor abstraction layer feeds data into the perception layer, which uses computer vision
techniques (including deep learning~\cite{redmon2016you}) to detect static  objects (e.g., lanes,
traffic signs, barriers) and dynamic objects (e.g., passenger vehicles, trucks, cyclists,
pedestrians) present in a driving scenario.

The object detection algorithm performs several tasks (e.g., segmentation, classification, and
clustering). It uses all the sensor data separately and then merges the data using sensor fusion
algorithms (e.g., extended Kalman filtering~\cite{reid1979algorithm, julier1997new}). The fusion
algorithm provides software-level data redundancy for object detection. Use of HD maps and the
localization module enables the ADS to predetermine the location of specific static objects, such as
traffic lights, further improving the confidence in obstacle-detection tasks.

The perception layer is also responsible for temporal tracking of objects and lanes. Tracking is
necessary to ensure that an object does not suddenly disappear from a frame because of
misclassification or a failure to detect anything.
Thus, sensor fusion and tracking provide spatial and temporal redundancy in the perception layer of
the software.  After accurate determination and tracking of objects and lanes are completed, the
perception layer calculates various useful metrics such as ``closest in path obstacle'' (CIPO) and
``tailgating distance'' for each object. Such association of an object with measured or inferred
metrics (e.g., CIPO and tailgating distance) is defined as the \emph{world model} of the AV.

\textbf{Localization Layer} (\circled{3} in \cref{fig:appollo-arch}):
The localization module is responsible for aggregating data from various sources to locate the
autonomous vehicle in the world model. Localization in the world model can be done using a GPS
sensor or by using camera/LiDAR inputs. The work described in this paper uses only camera/LiDAR
along with maps to enable localization (i.e., it does not use GPS).

\textbf{Prediction Layer} (\circled{4} in \cref{fig:appollo-arch}):
The prediction layer is responsible for generating trajectories for detected objects by using
information from the world model (e.g., positions, headings, velocities, accelerations). As a
result, it can probabilistically identify obstacles in an AV's path~\cite{houenou2013vehicle}.

\textbf{Planning \& Control Layer} (\circled{5} in \cref{fig:appollo-arch}): 
The planning and control layer is responsible for generating navigation plans based on the origin
and destination of the EV and for sending control signals (actuation, brake, steer) to the AV.
The ``Routing module'' generates high-level navigation information based on requests. The Routing
module needs to know the routing start point and routing end point, in order to compute the passage
lanes and roads. The ``Planning module'' plans a safe and collision-free trajectory by using
localization output, prediction output, and routing output. The ``Control module'' takes the planned
trajectory as input and generates the control command to pass to the CAN Bus, which passes the
information to the AV's mechanical components.
The surveillance system monitors all the modules in the vehicle, including hardware. The ``Monitor
module'' receives data from different modules and passes them on to a human-machine interface for
the human driver to view to ensure that all the modules are operating normally. In the event of a
module or hardware failure, the monitor triggers an alert in the ``Guardian module,'' which then
chooses an action to be taken to prevent an accident.

\subsection{Simulation Platform} \label{sec:sim_platform}
\begin{figure}[!t]
    \centering
    \includegraphics[width=\columnwidth]{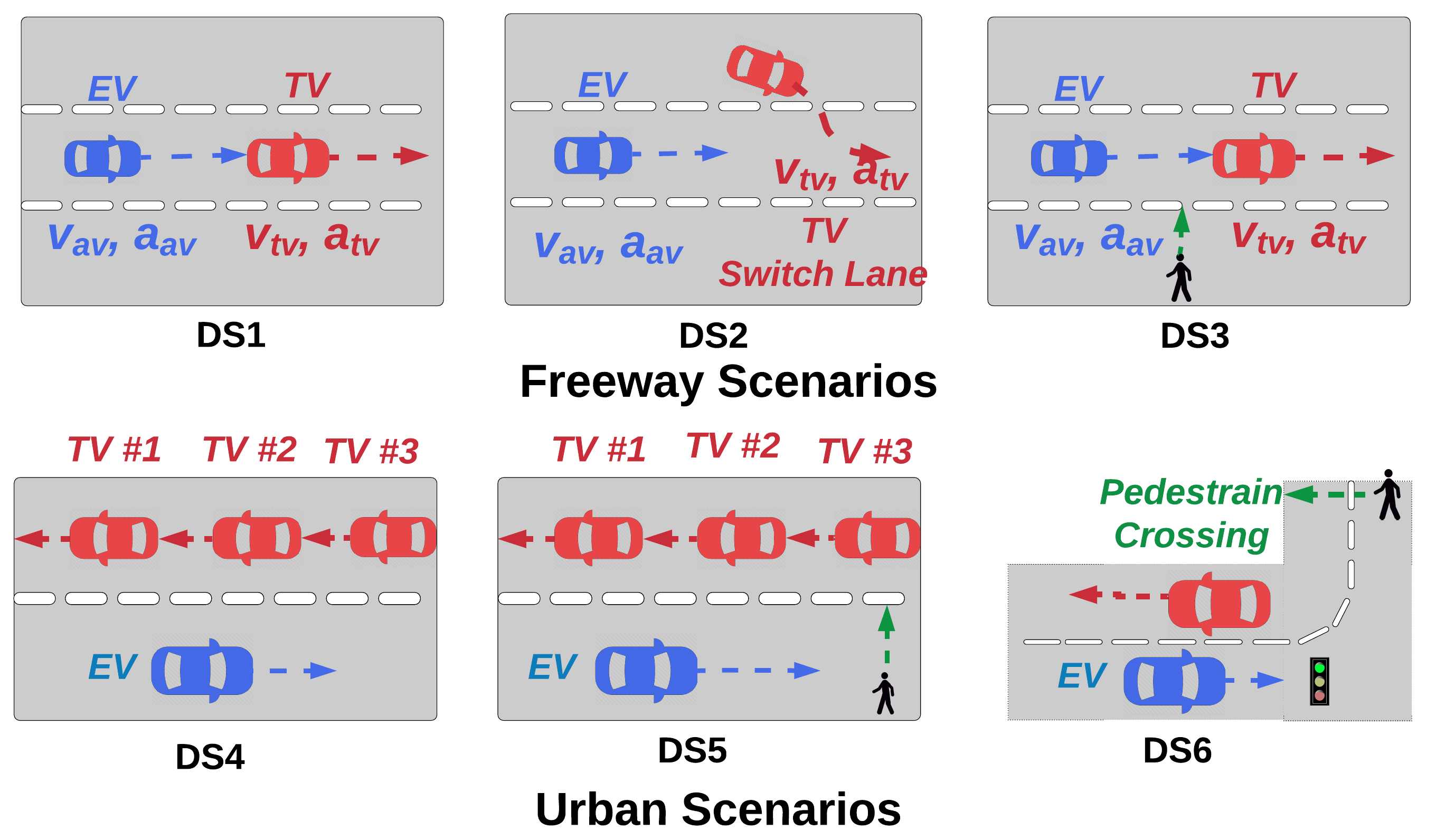}
    \caption{Driving scenarios supported by simulation engine.}
    \label{fig:driving-scenarios}
    \spacefigure
\end{figure}

This paper uses Unreal Engine (UE) based simulation platforms (Carla~\cite{Dosovitskiy17} and
DriveSim~\cite{drivesim})  that are capable of simulating complex urban and freeway driving scenarios
by using a library of urban layouts, buildings, pedestrians, vehicles, and weather conditions (e.g.,
sunny, rainy, and foggy). The simulation platforms are capable of generating sensor data at regular intervals
 (from cameras and LiDARs) that can be fed to the ADS platform. A driving scenario
consists of 500 scenes in \dw or 2400 scenes in \apollo in which the EV travels from a fixed
starting point on the road to a fixed destination point. A scene in a driving scenario is a
representation of the physical world at the simulation epoch and corresponds to a camera frame.
Fig.~\ref{fig:driving-scenarios} illustrates scenes from three freeway (DS1--DS3) and three urban (DS4--DS6)
driving scenarios used in this study. DS1--3 are controlled by \dw in DriveSim, and DS4--6 by \apollo
in Carla. In these scenarios an EV and a few UE-controlled TVs/pedestrians are
placed in urban and freeway roads, driving at different velocities/accelerations and separated by some
distance. The EV is expected to execute driving maneuvers in each of these settings. The scenarios
represent the most common driving cases encountered by humans on a daily basis. In DS1--DS6, the Ego
vehicle does not switch lanes, there is no other vehicle trailing the Ego vehicle, and the Ego
vehicle is in a safe state.
\subsection{Hardware Platform}
The NVIDIA DriveAV ADS was designed for the NVIDIA AGX Pegasus platform~\cite{pegasus}, which
consists of two Xavier SoCs and two discrete GPUs, but is also supported on a development platform
based on an x86 CPU and a GPU. For our experiments, we used the development platform and its
utilities to facilitate the creation of the DriveFI tool. The Apollo ADS is supported on the
Nuvo-6108GC~\cite{nuvo6180}, which consists of Intel Xeon CPUs and NVIDIA GPUs.  We use Apollo on an
x86 workstation with two NVIDIA Titan Xp GPUs. %

%% file: 05000-drivefi.tex
\section{DriveFI Architecture}\label{sec:drivefi}
\begin{figure}
    \centering
    \includegraphics[]{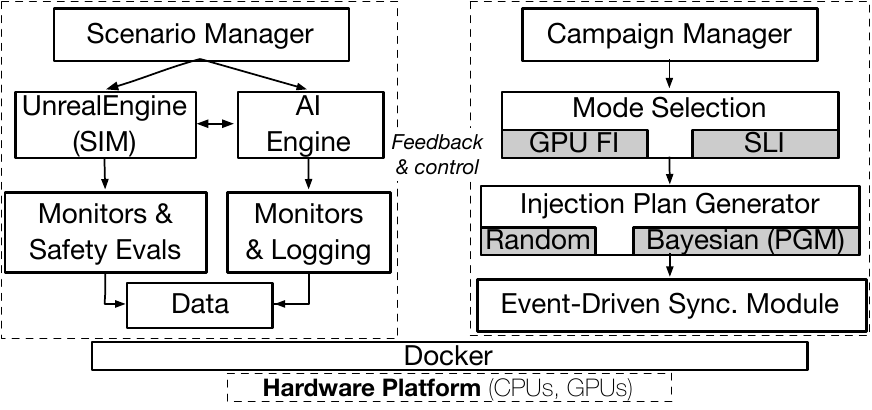}
    \caption{DriveFI architecture.}
    \label{fig:drivefi}
    \spacefigure
\end{figure}

The software architecture of \toolx is shown in ~\cref{fig:drivefi}. DriveFI leverages the existing tools to simulate driving scenarios and control the EV in simulation by using an AI agent (which is provided by \apollo or \dw). The scenario manager coordinates the simulator and AI agent to run a driving scenario and monitor the state of the software as well as the safety of the EV. \toolx is bundled with a campaign manager that takes an XML configuration file as input to select a fault model, software or hardware module sites for FI, the number of faults, and a driving scenario. The campaign manager uses the specified configuration to (a)  profile the ADS workload, (b)  generate a fault plan\footnote{A fault plan specifies which instruction/variable to corrupt, the corruption time, and the corruption value.}, and (c)  inject one or more transient faults per run into the ADS system. Based on the values in the configuration file, the campaign manager runs a specified number of golden simulations, profiles the ADS while running a driving scenario, and runs a specified number of experiments that inject one or more faults at a time based on the generated fault plan. %
The ``Event-driven synchronization'' module helps coordinate among all the toolkits (the UE-based driving scenario simulator, monitoring agents, campaign manager, fault injectors, and AI agent).

We built \toolx to characterize error propagation and masking (a) in computational elements, (b) in the ADS, and (c) in vehicle dynamics and traffic.  Low-level circuit, micro-architectural, and RTL faults manifest as architectural-state faults in computational elements.  The architectural-state faults that do not get masked manifest as errors in the internal state of the ADS modules, and the errors that do not get masked in the module propagate to the output of the module. Finally, errors that are not masked in any of the modules manifest as actuation command errors that are sent to the AV. Therefore, to mimic faults and errors, we built two fault injectors: (a) a GPU fault injector (GI; see Section~\ref{subsec:gpu_fm}) capable of injecting faults into the GPU architecture state to reveal the propagation of GPU faults to the ADS state, and (b) a source-level fault injector (SLI, see Section~\ref{subsec:sli_fm}) capable of injecting faults to corrupt ADS software variables. Corruption of the final output (actuation values $\zeta, b, \theta$) of the ADS helps us to measure the resilience associated with vehicle dynamics and traffic.  Thus, our approach aids in the measurement of fault and error masking/propagation at different levels and the corresponding impact on the safety of the AV.

\toolx is bundled with a campaign manager that takes an XML configuration file as input to select a fault model, software or hardware module sites for FI, the number of faults, and a driving scenario. The campaign manager uses the specified configuration (a) to profile the ADS workload, (b) to generate a fault plan, and (c) to inject one or more transient faults per run into the ADS system. %
For this paper, we developed an ``event-driven synchronization'' module that coordinates among all the toolkits (the UE-based driving scenario simulator, monitoring agents, campaign manager, fault injectors, and AI agent). %

\subsection{Injecting into Computational Elements: GPU Fault Models}\label{subsec:gpu_fm}
We consider transient faults in the functional units (e.g., arithmetic and logic units, and load store units), latches, and unprotected SRAM structures of the GPU processor. Such transient faults are modeled by injecting bit-flips (single and double) in the outputs of executing instructions. If the destination register is a general-purpose register or a condition code, one or two bits are randomly selected to be flipped. For  store  instructions,  we  flip  a  randomly
selected bit (or bits) in the stored value. Since  we  inject  faults  directly  into the live  state  (destination
registers), our fault model does not account for various masking  factors  in  the  lower  layers  of  the  hardware  stack, such
as circuit-, gate-, and micro-architecture-level masking, as well
as masking due to faults in architecturally untouched values. The GI employs an approach similar to that of SASSIFI~\cite{hari2017sassifi} and includes a profiling pass and fault-injection plan generation.
We do not consider faults in cache, memory, and register files, as they are protected by ECC.

\begin{table}[!t]
    \centering
    \caption{Examples of SLI-supported ADS module outputs.} \label{tab:sli_fm}
    \resizebox{\columnwidth}{!}{%
    \begin{tabular}{>{\ttfamily}p{\columnwidth}}
        \toprule
        \multicolumn{1}{l}{\textbf{FI Target (Output Variables)}} \\
        \midrule
        
        \rowcolor{gray!15}
        \multicolumn{1}{c}{\textbf{Path Perception Module}}\\
        lane\_type, lane\_width  \\
        
        \rowcolor{gray!15}
        \multicolumn{1}{c}{\textbf{Object Perception Module}}\\
        camera\_object\_distance, camera\_object\_class,
        lidar\_object\_distance, lidar\_object\_class, sensor\_fused\_obstacle\_distance, sensor\_fused\_obstacle\_class   \\
        \rowcolor{gray!15}
        \multicolumn{1}{c}{\textbf{Planning \& Control Module}}\\
        vehicle\_state\_measurements ($pos,v,a$),
        obstacle\_state\_measurements ($pos,v,a$),
        actuator\_values ($\zeta,b,\phi$), pid\_measured\_value, pid\_output \\
        
        \bottomrule
    \end{tabular}%
    }
    \spacefigure
\end{table}

\subsection{Injecting Faults into ADS Module Output Variables}\label{subsec:sli_fm}
The goal of SLI (Source-Level Injection) is to corrupt the internal state of the ADS by modifying ADS module output variables (hence, the input variables of another module) of the ADS components. SLI is implemented as a library that is statically linked to the ADS software; however, its use requires source-code modification and recompilation of the ADS software. We did not observe any noticeable runtime difference between SLI-linked ADS and non-SLI ADS. In this work, we manually identified the software variables that store the outputs of ADS modules that play a critical role in inferring the actuation commands of the EV. Source-code modification is required in order to mark the output variable and invoke the corresponding module injector to get a corrupted value by using the fault model provided in the XML config file. In Table~\ref{tab:sli_fm}, we show some of the variables from each of the ADS modules (see Fig.~\ref{fig:appollo-arch}) that were targeted using SLI.

The fault models supported by SLI that corrupt one or more software output variables in the $k$\textsuperscript{th} scene (chosen uniformly and randomly over all scenes of a driving scenario) are specified by (a) a number of faults (i.e., a number of consecutive scenes to be injected), and (b) the fault location. A \emph{single fault} in SLI-based experiments is the corruption of a single output variable of an ADS module. In the following, we define these SLI-supported fault models.%

\textbf{1-Fixed}. A single fault is injected at the $k^{\text{th}}$ scene of a given ADS software module output.  Across experiments, a constant value is used to corrupt the given ADS software module output. There are a total of 41 ``1-Fixed'' fault types, each defined by (a) the ADS module output, and (b) the corruption value. The bounded continuous outputs are corrupted to maximum or minimum possible value for those outputs, For example, to inject into brake actuation output, SLI uses a maximum brake value of 1.0 or a minimum brake value of 0.0. Unbounded continuous output values (e.g., $v$, $a$, and $pos$) are corrupted to double or half of the current output value\footnote{We limit ourselves to corruption of the outputs to double or half, as otherwise the ADS may detect the injected faults as errors.}. For categorical output variables the output value is corrupted to one of the categorical values; e.g., the  object/obstacle class can be corrupted to ``do not care/disappear,'' ``pedestrian,'' ``vehicle,'' and ``cyclist.''

\textbf{M-Fixed}. $m$ faults are injected into a given set of ADS software module output starting at scene $k$, and continues to inject faults into the ADS software module output until scene $K+m$. $m$ is chosen uniformly and randomly between 10 and 100.  The range selected for $m$ is large enough to support study of a threshold value for a number of consecutive frames/scenes that must be injected to cause a hazardous situation.  Again, there are 41 ``M-Fixed'' fault types.

\textbf{1-Random}. A single fault is injected at the $k^{\text{th}}$ scene in a uniformly and randomly chosen set of ADS module output.
The injected fault value is also chosen uniformly and randomly from the range of values of the selected ADS module output. 

\textbf{M-Random}. $m$ faults are injected in a set of randomly chosen ADS software module output starting at scene $k$, and continues to inject faults in the ADS software module output until scene $K+M$. $m$ is chosen uniformly and randomly between 10 and 100. In this case, both the ADS module and the corruption value are selected uniformly and randomly.

%% file: 06000-results.tex
\begin{table}[!t]
\centering
\scriptsize
\caption{Fault injection experiments.}
\label{tab:experiments}
\begin{threeparttable}
\centering
\begin{tabular}{lllll}
\toprule
\textbf{Campaign}       & \textbf{Target module}         & \textbf{\#Faults/Experiment}      \\
\midrule
1-GPU-all        & All GPU kernels         & 1                 \\
1-RANDOM            & All software module outputs     & 1                              \\
1-Fixed\_throttle\_max  & Actuator - throttle        & 1                                  \\
1-Fixed\_brake\_max     & Actuator - brake           & 1                                  \\
1-Fixed\_Steer\_max     & Actuator - steer           & 1                               \\
1-Fixed\_obstacle\_rem  & Perception - obstacle disappear & 1                                 \\
1-Fixed\_obstacle\_dist & Perception - obstacle distance  & 1                            \\
1-Fixed\_lane\_rem      & Perception - lane disappear      & 1                             \\
M-Random            & All software module outputs     & 10--100                              \\
M-Fixed\_throttle\_max  & Actuator - throttle        & 10--100                                 \\
M-Fixed\_brake\_max     & Actuator - brake           & 10--100                              \\
M-Fixed\_Steer\_max     & Actuator - steer           & 10--100                            \\
M-Fixed\_obstacle\_rem  & Perception - obstacle disappear & 10--100                           \\
M-Fixed\_obstacle\_dist & Perception - obstacle distance  & 10--100                                  \\
M-Fixed\_lane\_rem      & Perception - lane disappear      & 10--100                              \\
1-PGM            & All software modules            & 1                             \\
\bottomrule
\end{tabular}
\end{threeparttable}
\spacefigure
\end{table}

\section{Results}
\label{sec:results}
In this section, we characterize the impact of fault and error injection on the safety of the EV. In our work, we use a UE-based simulator to study three freeway driving scenarios (DS1--DS3) and three urban driving scenarios (DS4--DS6). DS1--DS3 were controlled by \dw, whereas DS4--DS6 were controlled by \apollo. 
The safety of the EV at any given scene is verified by calculating the CIPO (the closest in path obstacle) and LK distance(lateral distance from the center of the lane). A safety hazard occurs when $d_{min} < 1.0$ m in the longitudinal direction, which corresponds to less than 1.0 m of minimum distance from CIPO, or when the EV crosses the Ego lane, which corresponds to a 0.80 m displacement from the center of the lane. Hence, the minimum CIPO distance (min-CIPO) and maximum LK distance (max-LK) across all scenes characterize the safety hazard for the entire simulation.%

\begin{figure*}[t!]
    \centering
    \subcaptionbox{\apollo min-CIPO~\label{subfig:apollo-min}}{
    \includegraphics[width=0.27\textwidth]{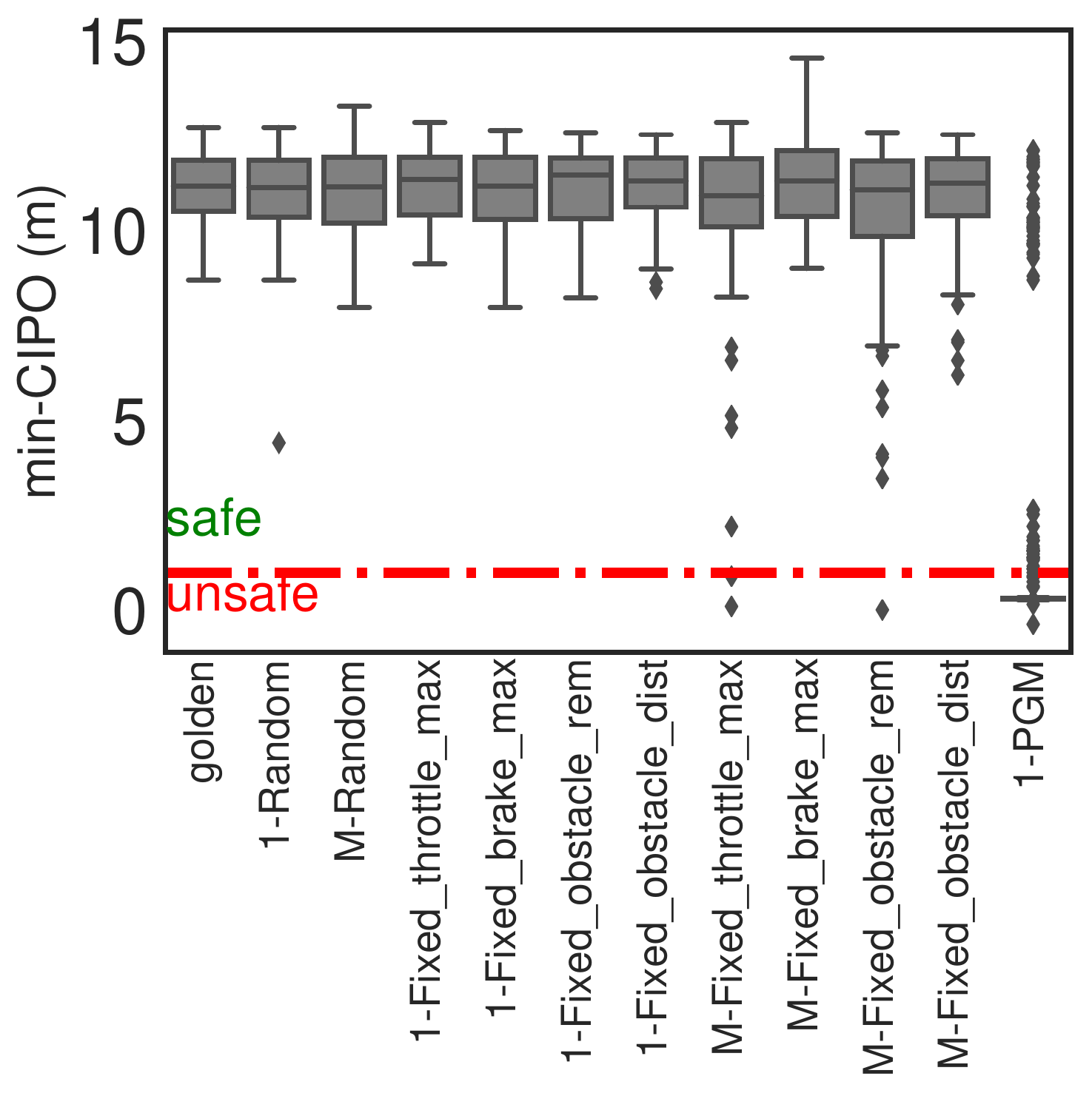}}
    \subcaptionbox{\apollo max-LK~\label{subfig:apollo-max}}{\includegraphics[width=0.18\textwidth]{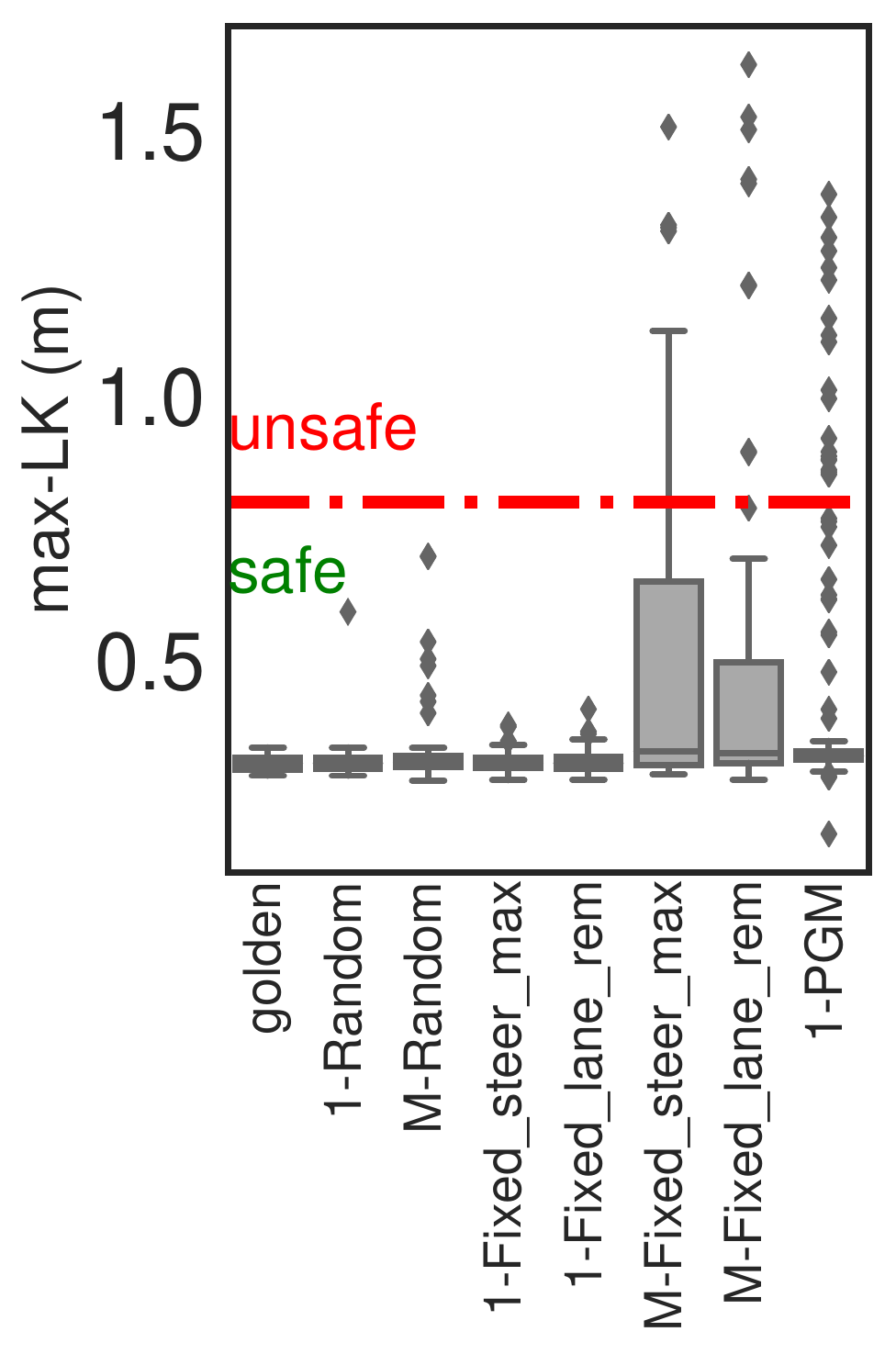}}
    \subcaptionbox{\dw min-CIPO~\label{subfig:dw-min}}{
    \includegraphics[width=0.27\textwidth]{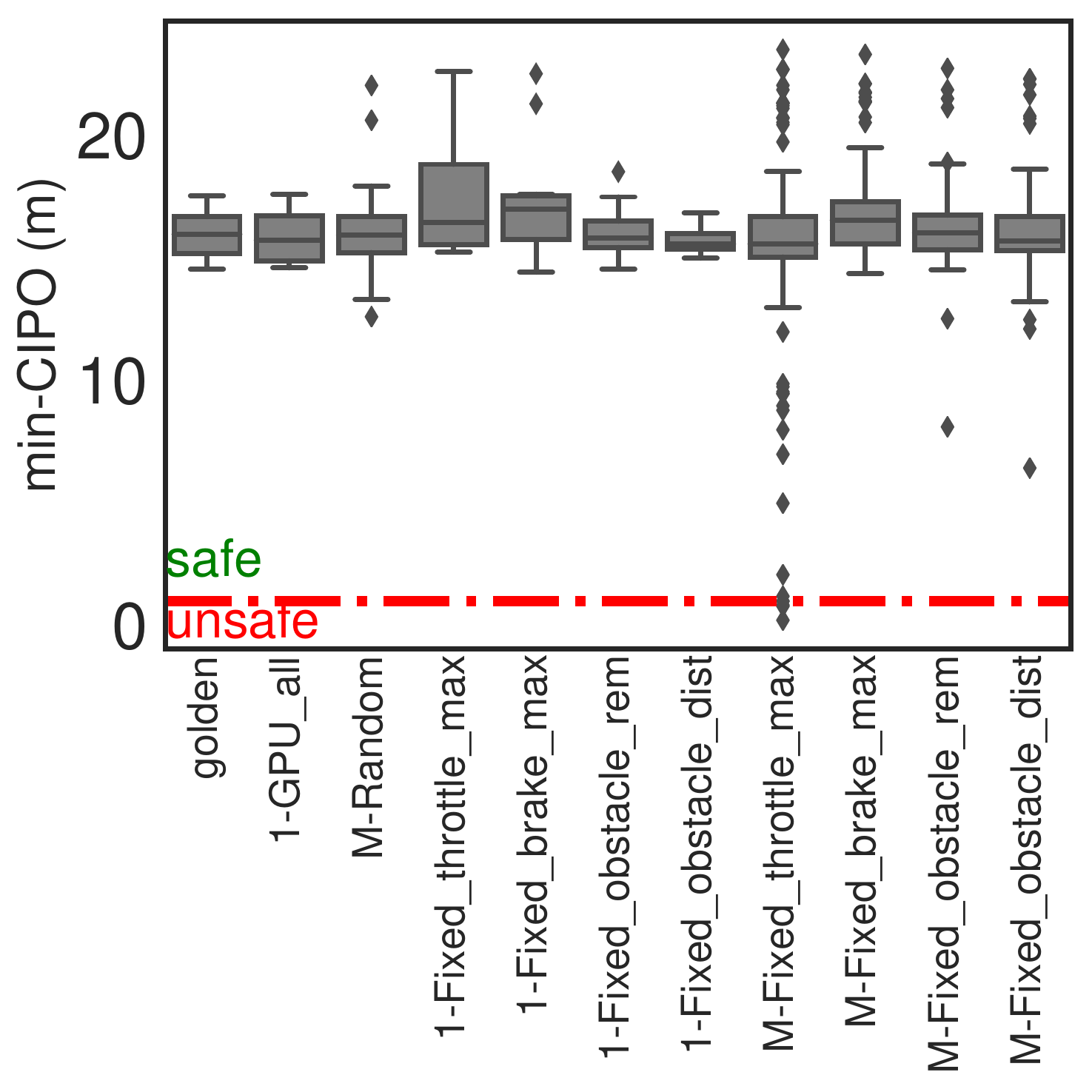}}
    \subcaptionbox{\dw max-LK~\label{subfig:dw-max}}{\includegraphics[width=0.18\textwidth]{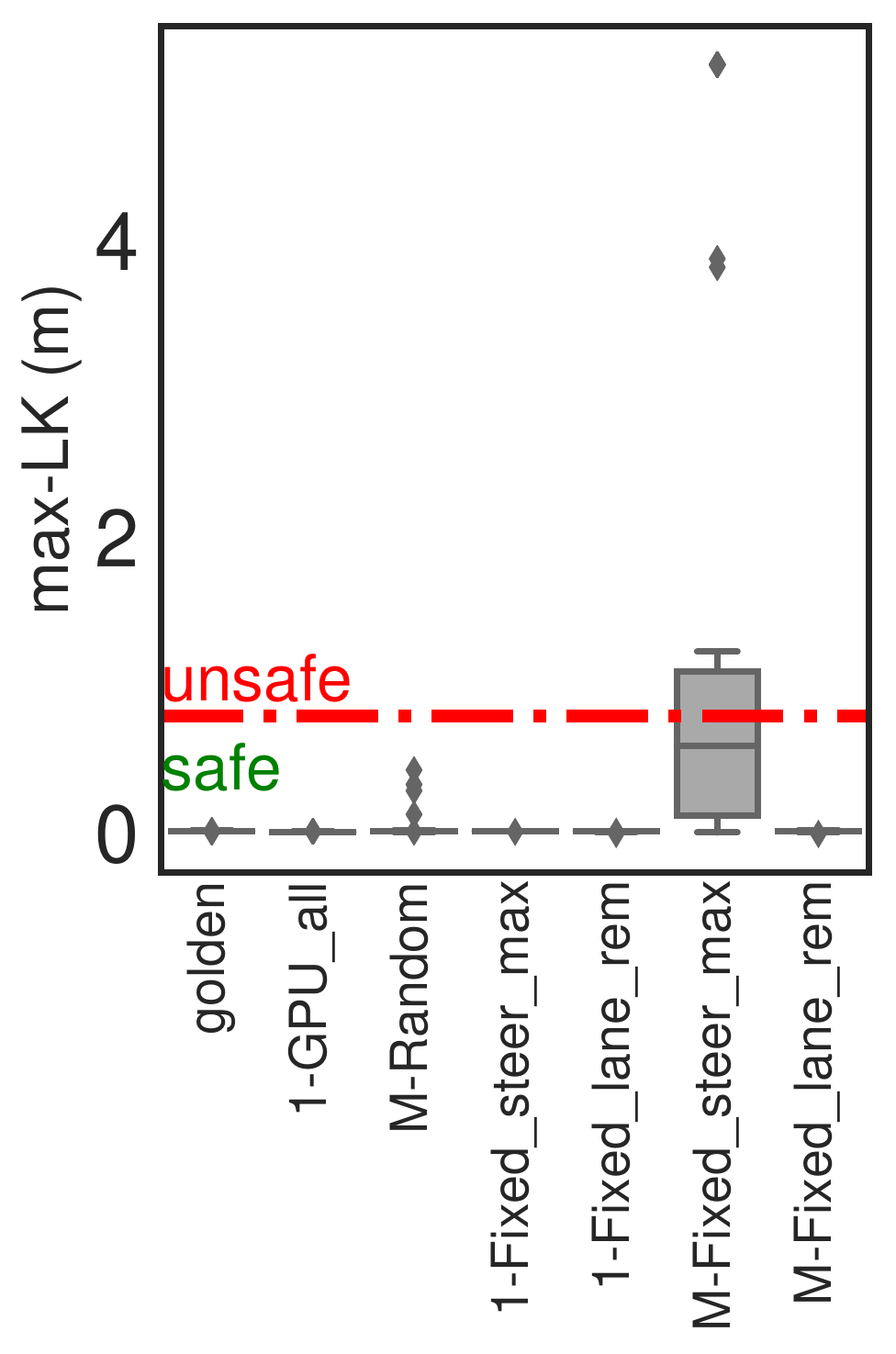}
    }
    \caption{Fault/error impact characterization using FI campaigns. (a) \& (b) use DS6; (c) \& (d) use DS1.}
    \label{fig:char}
    \spacefigure
\end{figure*}

Because of space restrictions, without any loss of generality, we limit our discussion to DS1, in which the EV was controlled by \dw, and DS6, in which the EV was controlled by \apollo.~\cref{subfig:apollo-min,subfig:apollo-max,subfig:dw-min,subfig:dw-max} show the boxplots of min-CIPO and max-LK for \apollo (DS6) and \dw (DS1), respectively, across all fault injection experiments and golden runs.  These experiments are summarized in \cref{tab:experiments}.  A  boxplot shows the distribution of quantitative data in a way that facilitates comparisons between variables or across levels of a categorical variable. The boxplot shows the quartiles of the dataset, while the whiskers extend to show the rest of the distribution (maximum and minimum samples), except for points that are determined to be outliers~\cite{boxplot}. To understand the simulation and safety characteristics of the driving scenarios, we ran 50 end-to-end simulations for each scenario without any injection. These runs are called \emph{golden runs}. The golden runs serve as a reference against which we compare injected simulation runs in the rest of the paper. The median min-CIPO and max-LK distances are 16 m (see ``golden'' in~\cref{subfig:dw-min}) and 0.019 m (see ``golden'' in~\cref{subfig:dw-max}) for \dw, and 11.19 m (see ``golden'' in~\cref{subfig:apollo-min}) and 0.31 m (see ``golden'' in~\cref{subfig:apollo-max}) for \apollo. None of the golden runs resulted in safety hazards.  

%% file: 06010-GPU-FI.tex
\subsection{GPU-level Fault Injection }~\label{subsec:gpu_results}
We conducted 800 GPU-level FI experiments for each driving scenario (DS1, DS2, DS3) in \dw. The min-CIPO and max-LK of DS1 simulated in \dw are labelled as ``1-GPU'' in \cref{subfig:dw-min} and \cref{subfig:dw-max}, respectively. We conducted only 800 GPU-level FI experiments per scenario because we did not observe any safety violations during the runs, and running more experiments would have been prohibitively expensive (2.7 days per driving scenario, 800*5 minutes/FI). In FI experiments labelled  ``1-GPU\_all'', faults were chosen uniformly randomly from across all dynamic instructions in the ADS. We did not conduct any GPU FI experiments on \apollo because of a CUDA driver version mismatch between GI and \apollo. Resolving the issue would have required vendor support and fixes. From \cref{subfig:dw-min} and \cref{subfig:dw-max}, we can observe that the EV is always safe, even after FI, and that the distribution is similar to the one in the golden case.

\textbf{Fault propagation and masking in GPUs.} Across all GPU-FI experiments on the DS1--DS3 driving scenarios, representing a total of 2400 FI experiments, 1.9\% of injected faults led to silent data corruption (i.e., caused corruption of actuation outputs which are the final outputs of the ADS module), and 0.02\% led to object misclassification errors\footnote{\emph{Object misclassification} refers to incorrect classification of an object, e.g., a pedestrian may be recognized as a vehicle.}. None of the object misclassification errors resulted in actuation output corruption. Our results indicate that the perception module (which is responsible for object detection and classification) is more resilient than other ADS modules. The reason is that the perception software takes advantage of sensor fusion (i.e., redundancy in sensing devices can compensate for a fault of a single sensor).  Across all driving scenarios, the SDCs did not result in any EV safety breach.

7.35\% of faults resulted in detectable uncorrectable errors (DUEs) that led to ADS software crashes (61\%) or hangs (39\%). The ADS is equipped to handle detectable errors and take corresponding corrective or safety measures. Although DUEs are more common than SDCs, it is expected that systems can recover from such faults via the backup/redundant systems.

\textbf{Errors persist for multiple frames}. In 2\%  of the misclassification error cases (recall that 0.02\% of GPU-level FIs led to misclassification errors), ADS perception module outputs were incorrectly classified for more than one frame, i.e., the impact of the injected fault persisted for more than one frame. In our data, we observed misclassification of objects for up to eight continuous frames. In those cases, errors did get masked eventually because of the temporal nature of the ADS platform. For example, ADS is fed with new sensor data at regular intervals, e.g., 7.5 times per second in our study.  
This observation suggests the need for more thorough study of fault masking and propagation in ADSs at the software level to handle cases in which faults persist for more than one frame.

%% file: 06100-SLI.tex
\subsection{Source-level Fault Injections}~\label{subsec:sli_results}
 We observed in the previous section that the ADS was able to compensate for injected transient faults. To further understand the ADS platform's susceptibility to faults and its robustness in the case of persistent errors, we conducted targeted FI with SLI to inject one or more faults directly into the ADS module  outputs. %
 We conducted 84 SLI-based FI campaigns for each driving scenario (scenarios 1--3 in \dw and 4--6 in \apollo). Of the 43 campaigns, 1 corresponded to ``1-Random,'' 1 corresponded to ``M-Random,'' 41 corresponded to 41 fault types under ``M-Fixed,'' and 41 corresponded to 41 fault types under ``M-Random.'' Labels are shown in \cref{fig:char}.   %

\textbf{Robustness of the ADS to single and multiple faults.}
The ADS platform was found to be robust to injection of a single fault (``1-Random'' campaign). To understand the robustness to persistence of fault-generating multiple random errors, we conducted FI campaigns on driving scenarios by using ``1-Random'' and ``M-Random'' fault models. The distributions of min-CIPO and max-LK for ``M-Random'' were found to be statistically different from those in the golden runs for \apollo (see ``M-Random'' in \cref{subfig:apollo-min} and \cref{subfig:apollo-max}) and \dw (see ``M-Random'' in \cref{subfig:dw-min} and \cref{subfig:dw-max}). For both ``1-Random'' and ``M-Random'' campaigns, none of the injected faults led to a hazardous driving situation; however, the ADS safety was found to be more vulnerable~\footnote{The AV came closer to the other vehicle/pedestrian compared to when no fault was injected.} to the ``M-Random'' fault model (especially for lane keep functionality). For example, the minimum min-CIPO observed across all injections decreased from 8.7 m to 8.0 m, and max-LK increased from 0.34 m to 0.7 m for \apollo. Similarly in \dw, min-CIPO increased from 15.2 m to 12.6 m, and max-LK decreased from 0.024 m to 0.43 m. 

\textbf{Robustness of the ADS modules to single and multiple faults.}
A persistent fault within the component of the ADS module continuously generates errors for the corresponding module. We tested the robustness of the ADS to a faulty module by subjecting one of the chosen module outputs to multiple faults. In these campaigns, we used ``1-Fixed'' and ``M-Fixed'' fault models. There are a total of 41 fault types for ``M-Fixed'' and ``1-Fixed'' fault types (e.g., ``throttle max,'' ``obstacle removal,'' and ``lane removal''). We discuss the results of only select campaigns because of lack of space.  The selected campaigns (shown in \cref{fig:char}) included (a) actuation module output corruption (in which the brake, throttle, and steering were all changed to the ``max'' allowed value); (b) sensor fusion output corruption (in which the obstacle class was changed to “disappear” and the distance that could be considered in trajectory planning was changed to “max”); and (c) lane output corruption (in which the lane type was changed to “disappear”). The FI experiments that led to safety breaches appear as data points below the red line for min-CIPO and above the red line for max-LK. Clearly, none of the FI campaigns conducted under the ``1-Fixed'' fault model led to safety hazards, but few were observed for ``M-random'' FI campaigns. We rank ADS modules by their module vulnerability factor (MVF), which we calculate by finding the percent of simulations that resulted in either (a) a min-CIPO distance less than the  minimum min-CIPO distance across the golden runs, or (b) a max-LK distance maximum more than the max-LK distance across golden simulation runs. Using that method, we find that the ``steer angle'' (MVF=46\%), ``lane classification'' (MVF=43\%), ``obstacle classification'' (MVF=10\%), and ``throttle'' (MVF=7\%) are most vulnerable for \apollo, whereas for \dw we find the same components to be vulnerable except for ``lane classification'' and ``obstacle removal''. 

The higher resilience of ``lane classification'' and ``obstacle removal'' in \dw can be attributed to the free-space detection module (not present in \apollo) and the scene attributes.  The free-space detection module helps the \dw EV to detect drivable space (using a dedicated DNN network tasked with finding drivable space) even if the object is misclassified or its attributes (such as distance and velocity etc.) are corrupted. The free-space detection module ensures safety without requiring complete replication of obstacle  detection and classification modules.  The masking of faults in both modules can also be attributed to obstacle registration and tracking in the world model that helps track the obstacle over time.

\begin{figure}[!t]
\centering
\includegraphics[width=0.9\columnwidth]{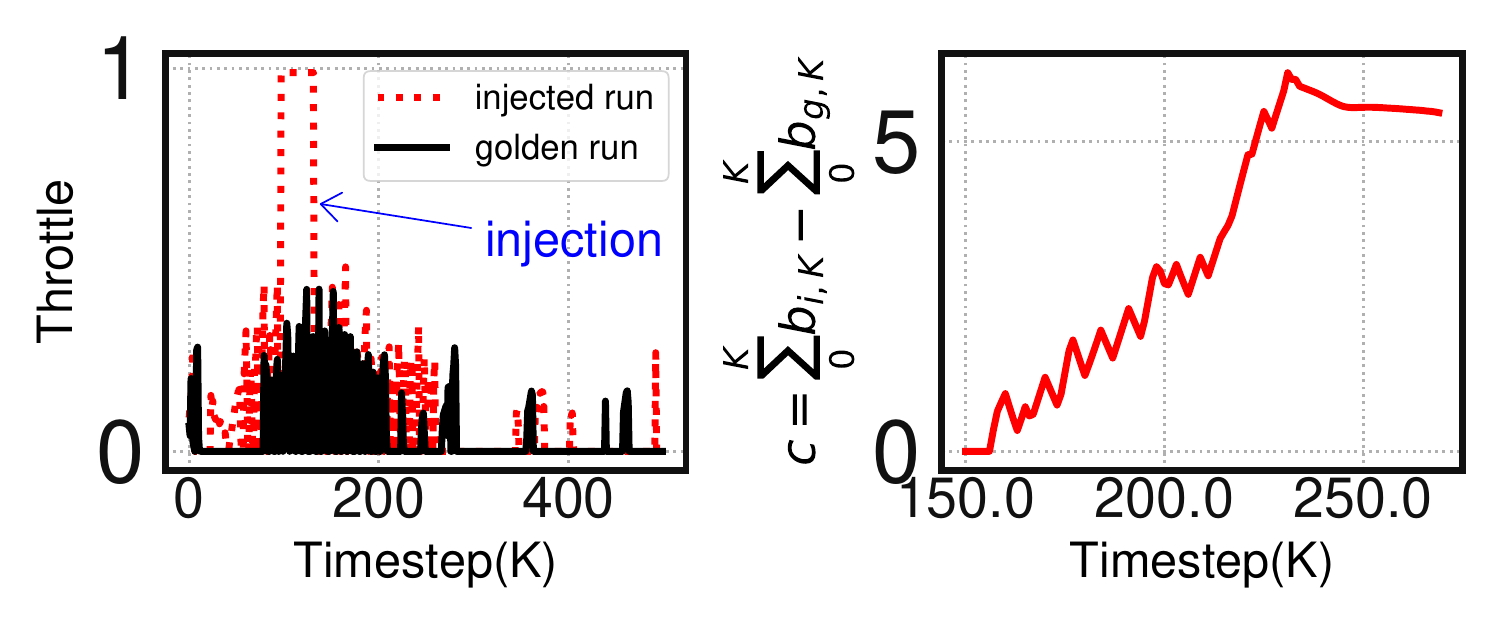}
\vspace{-2mm}
\caption{\small{Impact of 30 continuous faults on $\zeta$ in \dw. Left subfigure shows $\zeta$ for a golden simulation (in black) and an injected simulation (in red). Right subfigure shows compensation $c$.}}\label{fig:fi_greater}
\spacefigure
\end{figure}

\textbf{Compensation in ADS}: 
An ADS automatically compensates for any change in EV state (i.e., $ \theta, v, a, s$) that leads to an unsafe state caused by one or more faults/errors. It does so by issuing actuation commands that bring the EV to a safe state. For example, the EV may compensate for an increased $v$ by braking ($b$), a decreased $v$ by throttling ($\zeta$), or a change in heading angle by steering ($\phi$). \cref{fig:fi_greater} shows throttle ($\zeta$) values for golden and injected runs (in the left subfigure) and compensation achieved by braking (in the right subfigure) for an FI experiment in which $\zeta$ was corrupted in 30 consecutive frames/scenes. Compensation at time step $K$ is calculated as the difference between the cumulative sums of ``brake'' values observed at time step $K$ in the injected run and in the golden runs. The injection leads to an increase in the velocity of the vehicle, which is compensated for by braking. In the right subfigure in \cref{fig:fi_greater}, we show that the compensation increases until time step $K=232$ to undo the effects of multiple faults, and then flattens out as the brake values in the golden run and faulty run (i.e., run with fault injection) become equal. We observed similar compensation behavior for the faults injected into brake and steer values. 

The ability of an ADS to compensate for injected faults depends on the number of faults and the time of injection. The outlier data point below the red line in \cref{subfig:apollo-min} for ``M-Fixed\_throttle\_max'' corresponds to 30 consecutive frames/scenes injected with faults into  $\zeta$ values. In this FI experiment (not shown in \cref{fig:fi_greater}), the vehicle was not able to compensate for the injected faults, as the faults were injected at $K=400$ and there was not sufficient time for the vehicle to stop, i.e., the EV reached an unsafe state at the end of the injections. In \apollo, only 20 injected faults into $\zeta$ values led to unsafe states. \textit{Persistent errors have significant impact on the EV's state, and the ADS's ability to compensate for the impact of errors depends on the time and location of FIs.}

%% file: 06200-pgm.tex
\subsection{Results of Bayesian FI-based injections}
In our FI campaigns thus far, hazardous driving conditions (accidents and lane violations) were created only when multiple faults had been injected into the ADS (i.e., multiple consecutive frames/scenes had been injected). However, in the real world, it is more likely that a single fault will occur, and therefore it is important to find conditions under which a single fault can lead to hazardous driving conditions. One way to approach the problem of finding all such single faults (i.e., critical faults) is to inject every single fault while running a driving scenario in a simulator. That approach, however, would be prohibitively costly and is infeasible in practice. For example, an exhaustive search to find which of the 41 fault types under the ``1-Fixed'' fault model will lead to safety hazards would have taken 272 days\footnote{$\nicefrac{\text{615 days}}{DS} = \nicefrac{\text{9 min}}{DS} * 41 \text{\xspace fault types} * 2400 \text{\xspace scenes}$.}~\footnote{Note that traditional FI is sampling-based, so 615 days represents the worst case of enumeration of all faults.} in our simulation platform .  Another way to find critical faults is to inject faults uniformly and randomly. However, the results from GPU hardware-level FI (see \cref{subsec:gpu_results}) and ADS software module-level FI (see \cref{subsec:sli_results}) suggest that we need a smart FI method capable of identifying hazardous situations in driving scenarios and using them to guide FI experiments.  A fault injector based on such a method would inject a fault when the ADS is most vulnerable (i.e., the fault is likely to propagate to actuators) and in such a way that the ADS cannot compensate for the fault. The Bayesian fault injector is able to find a \emph{critical situation} that was inherently safe (i.e.,
$\delta > 0$) but became unsafe after injection of fault $f$ (i.e., $\delta_{\Prdo(f)} \leq 0$). We have shown the effectiveness of Bayesian FI by injecting faults into driving scenarios DS4-DS6 controlled by \apollo.

\textbf{Effectiveness of Bayesian FI.} When we used Bayesian FI, 82\% of injected faults resulted in hazards. (95\% of the hazards were accidents involving a pedestrian, and 5\% were lane violations.) Bayesian FI selects one of the 41 fault types of the ``1-Fixed'' fault model, and uses SLI to inject a single fault into an ADS module output variable. Recall that in the``1-Fixed' fault model, the fault location (i.e., the ADS module output variable) and corruption value are defined by the fault type. In comparison, none of the random single FIs led to safety hazards.   The Bayesian FI results are marked as ``1-PGM'' in \cref{subfig:apollo-min} and \cref{subfig:apollo-max}. All data points below the red line in \cref{subfig:apollo-min} correspond to collisions,  and all data points above the red line in \cref{subfig:apollo-max}  correspond to lane violations. The median min-CIPO distance was 0.32 m, which is significantly less than the 11.19 m median value for golden runs. Although the median max-LK value did not change for the ``1-PGM'' campaign compared to golden runs, 5\% of the hazards were due to lane violations.

\textbf{Mining critical faults and critical scenes.}
As discussed before, injection of all fault types under the ``1-Fixed'' fault model of SLI would be prohibitively expensive. Bayesian FI helped us find all critical faults $|F_{crit}|$ for every scene and mine driving scenes that are more susceptible to faults. The critical faults mined by Bayesian FI can help designers understand the weaknesses of the system and corner cases under which a fault may lead to hazards, whereas the critical scenes can be used by designers to inject random faults (using GI or SLI) only in those scenes to help them understand the architecture vulnerability factor (AVF). We believe that the mining of critical scenes by Bayesian FI will have wider applicability beyond
our FIs here. Combination of results from a range of FI experiments to create a
library of scenes will help manufacturers develop rules and conditions for AV testing and safe driving. \cref{tab:pgm-summary} gives summary statistics of mined critical faults and scenes in the driving scenarios (DS4--6). A total of 561 faults were found to be critical across DS4--6. Upon inspecting the mined critical faults, we found that the top 3 most susceptible ADS module outputs for vehicle collision are the throttle value (24\% of 561 critical faults), the PID controller input (18\%), and the sensor-fusion obstacle class value (15\% of 561 critical faults). ADS module outputs targeted by Bayesian FI for creating lane violations are the (a) lane type value (2\% of 561), (b) throttle (1.4\%), and (c) steer (1.4\%).  56\% of the fault types were never used by Bayesian FI; for example, Bayesian FI never injected into the output of camera-sensor object classification module. 

For DS4, we did not find any critical scene or error. That was expected, as there was no trailing or leading vehicle around the EV in our driving scenarios. All the vehicles were in the other lane following a completely different trajectory, and one fault in this case would not be sufficient to make the EV cross into the adjacent lane. For DS5, 0.88\% of the scenes and 0.20\% of the faults were found to be critical. The critical scenes in this case correspond to a scene in which (a) the object (i.e., pedestrian) is first registered into the world model and (b) the EV then starts braking. In case of (a), the Bayesian FI chooses to remove the obstacle (e.g., by removing the obstacle, or misclassifying the object), and in the case of (b), the Bayesian FI chooses to accelerate the vehicle (e.g., by corrupting PID outputs or planner outputs). For DS6, we observed that 1.96\% of the scenes and 0.36\% of the faults were critical. We made a similar observation for DS5.  However, in addition, we found the EV to be susceptible to faults around turns. Bayesian FI in those cases chooses faults that correspond to a disappearing lane or steering value corruptions. The EV tends to follow the lead vehicle when the lane markings are missing.  However, in turns for which there is no lead vehicle to follow, such errors become critical. \textit{It is worthwhile to note that Bayesian FI was able to mine critical faults and scenes in 4 hours, and took approximately 54 hours to simulate all the extracted faults in the simulator.}

\begin{table}[!t]
    \centering
    \begin{threeparttable}
    \centering
    \caption{Summary of PGM-based fault injection.}
    \label{tab:pgm-summary}
    \begin{tabular}{llll}
    \toprule
    \multicolumn{1}{l}{\textbf{Driving scenario}} & \textbf{Crit. scenes \%} & \textbf{Crit. faults \%} & \textbf{Hazard rate} \\
    \midrule
    DS4 (2400 scenes) & 0    & 0.0  &  0.0\\
    DS5 (2400 scenes) & 0.88 & 0.20 &  0.36\\
    DS6 (2400 scenes) & 1.96 & 0.36 &  0.20 \\
    \bottomrule
    \end{tabular}%
    \begin{tablenotes}
            \scriptsize
            \item [1] Total faults (TF) in the ``FIXED'' fault model = \#scenes/DS * \#error types = 98400/DS
            \item [2] Critical scenes \% = \#scenes in which critical faults were found by \#scenes/DS
            \item [3] Critical faults \% = (Critical faults mined by Bayesian FI)/TF
        \end{tablenotes}
    \end{threeparttable}
    \spacefigure
\end{table}

%% file: 08000-related.tex
\section{Related Work}\label{sec:related}

AV research has traditionally focused on improvement ML/AI techniques.  However, as models are deployed at large scale on computing platforms, the focus changes to assessment of the resilience and safety features of the compute stack that drives the AV. Assessment of the safety and resilience of AVs requires robust testing techniques that are scalable and directly applicable in real-world driving scenarios. It is not scalable or practical to base a safety argument solely on statistical measures such as a billion miles on roads, or on simulations done on platforms such as CARLA~\cite{Dosovitskiy17} or Open Pilot~\cite{openpilot}, ~\cite{Anderson2016, Kalra2016}. Testing the robustness of an ADS has proven to be challenging and mostly ad hoc or experience-based~\cite{fraade2018measuring}. In particular, to test the functionality and design of the hardware and software components of an ADS, current methods rely on injection of invalid or perturbed inputs~\cite{Pei2017, rubaiyat2018experimental, jha2018avfi} or faults and errors~\cite{jha2018avfi, Li2017, jhakayotee} into an ADS in simulation or ADS components, and accrual of millions of miles on roads~\cite{Google_Safety}. 
 
However, these methods are not scalable because (a) they lack simulated or real datasets that would represent all kinds of driving scenarios~\cite{Anderson2016}; (b) it would take billions of miles of driving to add functionality or do a bug fix, in order to drive statistical measures~\cite{koopman2018toward}; (c) they are restricted to DNNs\cite{lu2017no, pei2017towards, Kalra2016, lakkaraju2017identifying, Li2017} and sensors~\cite{jha2018avfi, rubaiyat2018experimental}, even when DNNs form only a small part of the whole ecosystem; and (d) once the easy bugs have been fixed, finding rare hazardous events would be exponentially more expensive, as faults might manifest only under specific conditions (e.g., a certain software state).

%% file: 09000-conclusions.tex
\section{Conclusion}\label{sec:conclusion}

In this work we present \toolx, a fault injection tool, along with methodologies to empirically assess the fault propagation, resilience, and safety characteristics of the ADS, as well as to generate and test corner-case failure conditions. \toolx incorporates Bayesian and traditional FI frameworks which work in tandem to accelerate finding of the safety-critical faults.

%% file: ack.tex
\section*{Acknowledgments} \addcontentsline{toc}{section}{Acknowledgment}
This material is based upon work supported by the National
Science Foundation (NSF) under Grant No. CNS 15-45069.
We thank K. Atchley,
J. Applequist, K. Saboo, and S. Cui. 
We would also like to thank NVIDIA Corportation for software access and equipment donation.
Any opinions, findings, and conclusions or recommendations expressed in this material
are those of the authors and do not necessarily reflect the views of the NSF and NVIDIA.